\documentclass[journal]{IEEEtran}
\usepackage{cite}
\usepackage{amsmath,amssymb,amsfonts}
\usepackage{caption}
\usepackage{floatrow}   
\usepackage{subcaption}
\usepackage{algorithmic}
\usepackage{graphicx}
\usepackage{textcomp}
\usepackage{xcolor}
\usepackage{svg}
\usepackage{multicol}
\usepackage{multirow}
\ifCLASSINFOpdf
\usepackage{ragged2e}
\usepackage{booktabs}
\usepackage{verbatim}
\usepackage{float}
\usepackage{hyperref}
\usepackage{todonotes}
\usepackage{enumitem}
\usepackage{tikz}

\else

\fi

\begin{document}

\makeatletter
\newcommand{\printfnsymbol}[1]{%
  \textsuperscript{\@fnsymbol{#1}}%
}
\makeatother

%title ideas: 
\title{A Universal Metric for Robust Evaluation of Synthetic Tabular Data}

\author{Vikram S Chundawat\printfnsymbol{1}\thanks{$^*$Equal contribution.}, Ayush K Tarun\printfnsymbol{1}, Murari Mandal\printfnsymbol{2}\thanks{$^\dagger$ Corresponding author.}, Mukund Lahoti, Pratik Narang
\thanks{Vikram S Chundawat, Ayush K Tarun, Mukund Lahoti, and Pratik Narang are with Birla Institute of Technology and Science (BITS) Pilani, Rajasthan 333031, India (e-mail: \{f20180128, f20180258, mukund.lahoti, pratik.narang\}@pilani.bits-pilani.ac.in)}
\thanks{Murari Mandal is with School of Computer Engineering, Kalinga Institute of Industrial Technology (Deemed to be University) Bhubaneswar, Odisha 751024, India (e-mail: murari.mandalfcs@kiit.ac.in)}
}
\maketitle

\begin{abstract}
Synthetic tabular data generation becomes crucial when real data is limited, expensive to collect, or simply cannot be used due to privacy concerns. However, producing good quality synthetic data is challenging. Several probabilistic, statistical, generative adversarial networks (GANs), and variational auto-encoder (VAEs) based approaches have been presented for synthetic tabular data generation. Once generated, evaluating the quality of the synthetic data is quite challenging. Some of the traditional metrics have been used in the literature but there is lack of a common, robust, and single metric. This makes it difficult to properly compare the effectiveness of different synthetic tabular data generation methods. In this paper we propose a new universal metric, TabSynDex, for robust evaluation of synthetic data. The proposed metric assesses the similarity of synthetic data with real data through different component scores which evaluate the characteristics that are desirable for ``high quality'' synthetic data. Being a single score metric and having an implicit bound, TabSynDex can also be used to observe and evaluate the training of neural network based approaches. This would help in obtaining insights that was not possible earlier. We present several baseline models for comparative analysis of the proposed evaluation metric with existing generative models. We also give a comparative analysis between TabSynDex and existing synthetic tabular data evaluation metrics. This shows the effectiveness and universality of our metric over the existing metrics. Source Code: \url{https://github.com/vikram2000b/tabsyndex}
\end{abstract}

\IEEEpeerreviewmaketitle
\section{Impact Statement}
Tabular data synthesis plays an important role in ensuring privacy preservation in the data-driven systems. The data synthesis algorithms help in producing synthetic data which statistically resembles real data and can comply with the privacy protection regulations (such as European Union General Data Protection Regulation (GDPR), California Consumer Privacy Act) due to its synthetic nature. Measuring the quality of synthetically generated tabular data is a challenging task. The effectiveness of such \textit{metrics} will increase the confidence of both the user and the regulators about the privacy preservation of data. This paper presents a single score universal metric for evaluation of synthetic tabular data. The proposed TabSynDex metric in this paper ensures fast computation of loss while training, consistency in the comparative evaluation of different generative methods. Moreover, it puts a more stringent criteria to measure the closeness of the synthetic data to real data. The findings in this paper opens several future possibilities and highlights the need for further research on the development of better methods for tabular data synthesis.

\section{Introduction}
Artificial data synthesis can help in alleviating several issues that are associated with the collection, obtaining consent, distribution, and scarcity of real data. Real data is difficult to obtain in many cases such as personal data, medical data, case-sensitive data, etc. With the adoption of new data privacy regulations such as European Union General Data Protection
Regulation (GDPR)~\cite{voigt2017eu}, California Consumer Privacy Act (CCPA)~\cite{goldman2020introduction}), the access of private data is increasingly becoming more cumbersome. Sometimes, it may not be possible at all to collect large amount of real data. For example, in the case of medical data for some rare disease, the available data would be very limited. The objective in synthetic data generation is to produce samples that closely represent the real data. This enables learning and inferences for real-world applications without an explicit mapping to the real data. Such generated data can be used to train machine learning (ML) models for applications where real data is not abundantly available due to any reason such as stated above.
%and and  Privacy is also a critical factor.

%In the literature, several methods have been presented for synthetic data generation. Traditionally, the statistical approaches such as Bayesian networks~\cite{BayesianGen,kaur2021application}, Bayesian mixture model~\cite{hu2018dirichlet,deyoreo2020bayesian}, Gibbs sampling~\cite{park2013perturbed}, and random forests~\cite{caiola2010random} have been used to synthesize data. Some researchers have explored adaptive sampling~\cite{haiboADASYN2008} and minority oversampling~\cite{ChawlaSMOTE2002,chawla2003smoteboost,guo2004learning} as well. More recently, the generative adversarial networks (GAN)~\cite{goodfellow2014generative} have become a preferred approach to generate synthetic data. One of the advantages of GAN based methods is their ability to generate completely new data, unlike the statistical methods which extrapolate over the existing data. The GAN-based methods along with Gaussian mixture models have been widely used for tabular data synthesis~\cite{tableGAN,DCGAN,xu2018synthesizing,xu2019modeling,medgan,JSD,baowaly2019synthesizing}.

%For example, medGAN~\cite{choi2017generating} and medBGAN~\cite{baowaly2019synthesizing} have been frequently use to synthesize health data. However, they are limited to binary and count variables, which represent a small subset of potential medical data types.

\subsection{Motivation}
Measuring the quality of synthetically generated tabular data is a challenging task. Unlike image data, qualitative evaluation through visual inspection is infeasible for tabular data. Qualitative evaluation with the help of insights from experts is a highly inefficient process. While most generative adversarial networks (GAN)-based and variation auto-encoder (VAE)-based tabular data generation methods~\cite{tableGAN,DCGAN,xu2018synthesizing,xu2019modeling,kingmaauto} use the machine learning efficacy as one of the evaluation metrics, the remaining metrics are almost never consistent among different research articles. For example,~\cite{xu2019modeling} uses a likelihood fitness metric,~\cite{JSD} uses Jensen-Shannon divergence and Wasserstein distance,~\cite{support_coverage} uses metrics like Kullback-Leibler (KL) divergence, pairwise correlation difference, log-cluster and support coverage. The lack of a uniform metric makes it difficult to appropriately compare the effectiveness of different generative methods. Further, without a bounded single score metric (like accuracy or F1-score), it is not possible to support the training of deep learning models for analysis. This is one of the reasons why the existing deep learning based approaches struggle to provide stable results or analysis. These methods are unable to offer clear answers to the questions like: How many number of epochs are enough to converge? If the model is learning at all after a certain number of epochs? Brenninkmeijer and Hille~\cite{brenninkmeijer2019synthetic} proposed an interpretable and bounded metric for evaluating synthetic data. They compare various statistics of real and fake data to generate different scores and then finally aggregate all of these scores to get a final single score metric. However, for certain scores they use correlation between the corresponding statistics of real and fake data to limit the range between [0, 1]. It means that even if the statistics corresponding to real and fake data for a particular component score are close but do not increase or decrease in the same order, the respective score will be very less. The consequence is that even if we evaluate the metrics for two different subsets of the same dataset, it shows very less similarity. This is contradictory as both the subsets belong to the same distribution. We have reported this observation through experiments in~Section~\ref{sec:experiments}. In order to alleviate this problem we propose a new metric in this paper. 

\subsection{Our Contribution}
We propose a new unified metric TabSynDex for robust evaluation of synthetic tabular data. The TabSynDex is interpretable and bounded $[0,1]$. We show that our single score based metric is sufficient to evaluate the generated synthetic data. The TabSynDex score encapsulates all the important characteristics essential to judge the quality of synthetic data. It includes comparison of basic statistics and machine learning efficacy with real data, fair representation of all categories of data and also, distinguishability from real data. We demonstrate its utility by evaluating the results of various synthetic tabular data generating techniques in the literature~\cite{tableGAN,DCGAN,xu2018synthesizing,xu2019modeling}. Apart from that, we also show how this can be used (as a loss computing metric) in the training of deep learning models. While using TabSynDex to observe the training, we are also able to uncover some very interesting aspects of the training process of tabular data generating GANs.

\section{Related Work}
\subsection{Traditional methods for tabular data synthesis}
The earlier works have proposed numerous sampling methods to generate minority class samples for imbalanced data.~The ADASYN~\cite{haiboADASYN2008} method use a weighted distribution of minority classes to adaptively generate samples.~The SMOTE~\cite{ChawlaSMOTE2002} method over-samples minority class examples by k-nearest neighbours. Furthermore, the Borderline-SMOTE~\cite{BSMOTE} method extend this work by over-sampling the minority examples only near the borderlines.~Calleja and Fuentes~\cite{DBOS} used a randomized weighted distance scheme to generate minority class data and Yue et al.~\cite{SDACR} used probabilistic fitting and k-means clustering to generate synthetic data for regression and classification. Garcia et al.~\cite{garcia2008use} proposed variants of SMOTE which took into account proximity and the spatial distribution of the examples. For generative modelling of relational databases, Synthetic Data Vault (SDV)~\cite{SDV} was proposed. The SDV use a multivariate modelling approach to model the data.~The EMERGE~\cite{EMERGE} (synthetic Electronic Medical Records Generator) method generate medical records to preserve privacy of original records. A software package "Synthea" that simulates lifespans of synthetic patients for medical data generation was presented in~\cite{SYNTHEA}.~Young et al.~\cite{BayesianGen} used Bayesian networks to generate synthetic data and subsequently DataSynthesizer~\cite{DataSynthesizer} used greedy Bayes algorithm to construct Bayesian networks that model correlated attributes.

\subsection{GANs and VAEs for tabular data synthesis}
An important application of synthetic data generation is to enhance privacy preservation in the ML based products. It can help in training an ML model even with limited set of real data. More specifically, limited data availability is quite a common problem in healthcare sector. The synthetic data generation methods have significant utility in healthcare related applications.~MedGAN\cite{medgan} is one of the first method to employ GANs for tabular data synthesis. The aim of MedGAN is to produce realistic synthetic patient records to alleviate the privacy concerns of the patients regarding data sharing. It uses an encoder-decoder setup to preprocess categorical data and generates high dimensional discrete variables.~An improvement over this method is presented in medBGAN~\cite{baowaly2019synthesizing} that generates more realistic synthetic EHR data. However, these methods~\cite{medgan,baowaly2019synthesizing} are vulnerable to possible privacy exposure in the form of attribute and identity leaks. Another limitation is that they can not handle mixture of categorical and continuous columns. This limitation makes them useful only for a small subset of potential medical data types. Park et al. proposed table-GAN~\cite{tableGAN} based on Deep Convolutional GAN (DCGAN)~\cite{DCGAN} to generate statistically similar data to the original ones. They show that the ML models trained with the synthetic data exhibit similar performance to the model trained on the real data.~Xu and Veeramachaneni proposed TGAN~\cite{xu2018synthesizing} to generate data sets with continuous and categorical columns. It uses a generator with a unidirectional Recurrent Neural Network architecture with LSTM cells. The discriminator is a simple fully connected network.~They also identify the problems neural networks suffer when working with non-Gaussian distributed input. Therefore, the authors~\cite{xu2018synthesizing} use a Gaussian Mixture Model (GMM) with fixed number of modes on each individual continuous column to alleviate some of the issues. Xu et al.~\cite{xu2019modeling} made some improvements over the original TGAN~\cite{xu2018synthesizing} while also enabling conditional generation and proposed CTGAN. They use a Variational GMM instead of a GMM with a fixed number of modes. The generator in CTGAN use a fully connected network instead of a sequential network used in TGAN~\cite{xu2018synthesizing}. A variational auto-encoder (VAE)~\cite{kingmaauto} was adapted for tabular data synthesis in~\cite{xu2019modeling}. The generator was built by directly using the original data. More recently, Zhao et al.~\cite{JSD} developed a conditional table GAN named CTAB-GAN to effectively model diverse data types including a mix of continuous and categorical variables. Other notable works in the literature include~\cite{zhao2022ctab,support_coverage,Chen2019TheVO,brenninkmeijer2019synthetic,rajabi2021tabfairgan,astolfigenerating}.

\subsection{Evaluation metrics for synthetic tabular data generation methods}
Most of the existing works use the machine learning model efficacy and a variety of different metrics to evaluate the generated synthetic data.~Xu et al.~\cite{xu2019modeling} used a likelihood fitness metric and machine learning efficacy. The likelihood fitness metric is only applicable for simulated data with a known probability distribution. For real world datasets where the distribution is unknown, this metric is not suitable.~Goncalves et al.~\cite{support_coverage} proposed a support coverage metric that measures whether the rarer classes have been properly represented in the synthetic data. They also use log cluster analysis, KL divergence and pairwise correlation distance.~The KL-divergence is usually applied on an individual column. This doesn't account for the inter-column statistic. Thus, it is incomplete without the correlation analysis based metric. Snoke at al.~\cite{pmse_expected} use basic statistical evaluation along with an expected value of pMSE~\cite{woo2009global} that measures how much the synthetic data is differentiable from the real data.~Zhao at al.~\cite{JSD} use Jensen-Shannon divergence and Wasserstein distance.~Similar to KL-divergence, both these metrics don't account for the inter-column statistics.~Buczak et al.~\cite{EMERGE} use expert insights onto the data to asses the quality of the generated data.~However, the domain expert inputs are not easily available and are usually costly.~Chen et al.~\cite{Chen2019TheVO} evaluate the data generated by Synthea~\cite{SYNTHEA} using several clinical quality measures. Bayesian network generators~\cite{BayesianGen} were evaluated using performance of logistic regression on synthetic data.~Recently,~Brenninkmeijer and Hille~\cite{brenninkmeijer2019synthetic} proposed a similarity score to capture the similarities between the different statistics of the real and synthesized data. They use the basic statistical properties (mean, median, and standard deviation) and other elements such as correlation matrices, PCA and machine learning efficacy for computing the similarity score. Researchers have used different types of metrics in the literature. We notice that there is a lack of a uniform evaluation metric or a uniform single score metric which captures the intricacies and robustness of the generated synthetic data.~Moreover, most of these metrics are unbounded and a comparative analysis between the existing works in the literature become difficult due to different metric adopted by different papers. We also observe that a unified metric to validate the tabular data generated from GANs and VAEs is missing in the literature. A single score metric which captures all the relevant information about the generated data would not only help in evaluating the output generated by the GANs, VAEs after training but also in assessing the training of these GANs and VAEs.

\section{TabSynDex}
\subsection{TabSynDex Metric}
\label{tabsyndex_subsection}
In order to measure the potency of the generated tabular data, we decided to focus on the quality of the data along with the statistical similarities. Therefore, in addition to the basic statistics, correlation score, and machine learning efficacy, we also compute the propensity mean squared error and support coverage. All these
scores combined, can effectively represent the quality of generated data. Each of the metrics is bounded with a minimum value of 0 and a maximum of value of 1. In case a negative score is encountered, it is truncated to 0. This is done because we do not want a negative score to penalise good performance in other metrics. Similarly, the upper level for the scores are limited to value 1. The final value is an average of all the five scores. Different weights can be assigned to individual scores but equal weights are assigned for a vanilla study.

\subsubsection{Basic Statistical Measures}
We aggregate three basic statistics of data, i.e., \textit{mean, median, and standard deviation}. We calculate the error for each column of the generated and real data using Eq.~\ref{eq1}.
\begin{equation}
\label{eq1}
    e = \frac{1}{n_{c}}\sum_{i = 1}^{n_{c}}\bigg| \frac{R_i - F_i}{R_i} \bigg|
\end{equation}
where $n_{c}$ is the total number of samples in a column, $R_i$ and $F_i$ are the statistical measures for a column of the real data and generated data, respectively. The relative errors are clipped to keep it at the maximum value of 1. This is done so that poor performance on a single column does not lead to an overall negative score. We also do not expect a good synthetic data generation technique to have an error of more than $100\%$ on basic statistics. The scores for a single metric like mean is calculated as
\begin{equation}
\label{eq2}
    s_{mean} = 1 - \frac{e_1^{mean}+e_2^{mean}+.....+e_n^{mean}}{n}
\end{equation}
where $n$ is the number of columns and $e_i^{mean}$ is the relative mean error for the $i^{th}$ column. The scores for median and standard deviation are calculated in a similar manner. Finally, the individual scores are aggregated as
\begin{equation}
\label{eq3}
    S_{basic} = \frac{s_{mean} + s_{median} + s_{std\_dev}}{3}
\end{equation}

\subsubsection{Log-transformed Correlation $S_{corr_l}$}
This statistical measure considers the correlation between the columns of each dataset. We deal with the categorical columns in the same way as in~\cite{brenninkmeijer2019synthetic}, i.e. we use Pearson’s correlation for association between two continuous columns, correlation ratio~\cite{corr_ratio} for association between categorical and continuous columns, and~\textit{Theil’s U}~\cite{theilU} for association between two categorical columns. This approach produces two matrices consisting of $n^2$ values in total, where $n$ is the number of columns in the datasets. These matrices are used to calculate the correlation scores. Similar to this, we calculate the relative errors for each entry in the correlation matrix of real data, clip it to a maximum value of 1, and subtract the average value from 1 to obtain the re-scaled score.\par

\textbf{Why log-transformation?:} One of the issues with directly calculating relative error is that the correlation matrix of real data can have very small values. These values may even be of the order $10^{-6}$, $10^{-5}$, and $10^{-4}$ whereas, the corresponding entries in correlation matrix for generated data can be of an order bigger or smaller.
This may not necessarily mean the generated data is bad, as both the correlations are nearly 0. But the relative error can result in a higher value. Therefore, the correlation score remains bad in almost all cases. To address this issue we use the log of the magnitudes of the entries of the association matrices while keeping the signs intact as given in Eq.~\ref{eq4}. 
\begin{equation}
\label{eq4}
    \hspace*{-1cm}   S_{corr_l} = 1 - \frac{1}{n^2-n}\sum_{i = 1}^{n}\sum_{\substack{j = 1\\ j\neq i}}^{n}\bigg| \frac{sln(r_{ij}) - sln(f_{ij})}{sln(r_{ij})} \bigg|
\end{equation}
where $sln(x) = sign(x)ln(|x|)$, $r_{ij}$, and $f_{ij}$ are corresponding entries of the association matrices of real and generated data. The $sign(x)$ equals 1 if $x$ is positive and -1 if $x$ is negative.

\subsubsection{$S_{pMSE}$ Index}
We introduce the $S_{pMSE}$ index to more robustly discriminate between the real and synthetic data. The $S_{pMSE}$ index is built upon the Propensity Mean Squared Error (\textit{pMSE}) which is commonly used in the literature to analyze the difference between the distributions of real data and synthetic data. If the real data contains $N_{real}$ samples and synthetic contains $N_{syn}$ samples. Then the combined data contains $N_{syn} \cup N_{real}$ samples where the label~\textit{Y} denotes real data with 0 and synthetic data with 1. We train a logistic regression model on the combined data for classification. In our analysis we only use the input variables and not their polynomial combinations as in~\cite{pmse_expected}. Because that would make computing the similarity score highly expensive when the number of columns are large. Thus, it would render the metric unsuitable for the purpose of training where the metric needs to be quickly calculated at each epoch. Furthermore, we use the regression model to predict the probabilities $\hat{p}_i$ of each sample being synthetic or real. The~\textit{pMSE} score is calculated as in Eq.~\ref{eq5}.
\begin{equation}
\label{eq5}
    pMSE = \frac{1}{N}\sum_{i=1}^{N}(\hat{p}_i - c)^2
\end{equation}
where $N$ is the total number of samples after combining the synthetic and real samples i.e., $N = N_{real}+N_{syn}$, $\hat{p}_i$ is the predicted probability of the $i^{th}$ sample being synthetic, and $c$ is the proportion of the samples being synthetic i.e., $c = N_{syn}/(N_{real}+N_{syn})$.\par

Snoke et al.~\cite{pmse_expected} introduced $pMSE_0$, the expected value of pMSE when the synthetic data is indistinguishable from the real data as given in Eq.~\ref{eq6}.
\begin{equation}
\label{eq6}
    E(pMSE_0) = \frac{(k-1)(1-c)^2c}{N}
\end{equation}
where $k$ is the number of parameters in the logistic regression model including the bias. The ratio between the $pMSE$ and the expected $pMSE_0$ is calculated to evaluate the generated data in~\cite{pmse_expected}. The more this ratio is close to 1, the fake data is more indistinguishable from the real data.\par 

We further improve this metric by introducing a factor $\alpha^{-|1-ratio|}$ to this score. This also standardizes the value of the ratio within the range of $[0,1]$. The value of $\alpha$ can be chosen according to the application and after deciding what pMSE value would be appropriate. For example, we observed that $\alpha=1.2$ works quite well when the synthetic data is in the range of $[0,5]$. The $S_{pMSE}$ index is defined as
\begin{equation}
\label{eq7}
    S_{pMSE} = \alpha^{|1-ratio|}
\end{equation}
where $ratio = pMSE/E(pMSE_0)$.

\subsubsection{Regularized Support Coverage $S_{cr}$}
We propose a support coverage~\cite{support_coverage} based metric for histogram comparison between the real and synthetic data. The support coverage measures the amount of variables support (available in the real data) covered in the synthetic data. We penalize the overall support coverage score the most if the rarer categories of a variable are not represented well. The metric is calculated as follows for a single variable or column $c$
\begin{equation}
\label{eq8}
    s_{cr} = \frac{1}{n_{cat}}\sum_{i=1}^{n_{cat}}\frac{n^f_i}{n^r_i}*scaling\,factor
\end{equation}

where $n_{cat}$ is the number of categories for a variable, $n^f_i$ is the number of samples in that category in fake data, $n^r_i$ is the corresponding number of samples in real data and the scaling factor is $N_{real}/N_{fake}$ i.e., the ratio of the number of samples in real data and the number of samples in fake data. For a categorical column, $n_{cat}$ is the number of categories. For continuous data, $n_{cat}$ is the number of bins of equal range that we divide the data into for this analysis. In our analysis we use 20 bins or $5\%$ range per bin. This ensures the bins are not so large as to disregard the distribution of values in the bins themselves, and not so small that there exist categories with no samples in them. Thus, we keep a balance so that this score fairly compares the distribution of values in the real and synthetic/fake columns.\par

This score for a single variable is affected the most if $n^r_i$ is small. It means the overall score may become much greater than 1 if the corresponding $n^f_i$ is even slightly bigger, and much smaller if it is even slightly lesser. We want to see a smaller score if a rarer category is not well represented, but not disregard the representation of other categories if a single rare category is too well represented. So we limit this score to a maximum of $\beta$, which can be decided based on the desired representation of categories. In our analysis we have used $\beta = 2$. The final score is an average of the scores for all the columns. The score of each column is limited to a maximum of 1. This is done so that a good coverage of representation of one column does not make the overall score disregard the bad coverage of other columns.

\begin{equation}
\label{eq9}
    S_{cr} = \frac{s_{cr1}+s_{cr2}+.......+s_{crn}}{n}
\end{equation}
where $s_{ci}$ is the score for the $i^{th}$ column, and $n$ is the number of columns.

\subsubsection{Machine Learning Efficacy based Metric}
We adopt the metric presented in~\cite{brenninkmeijer2019synthetic} i.e., we train a common model on both real and synthetic data. The trained model is evaluated on the real and synthetic test set. We calculate the relative error with respect to the RMSE score of a regression model trained on real data. Similarly, the f1-score of the classifier trained on real data is computed. It is to be noted that all the scores are scaled within $[0,1]$. The average of different errors is subtracted by 1 as given in Eq.~\ref{eq10}.

\begin{equation}
\label{eq10}
    S_{ml} = 1 - \frac{e_{m1}+e_{m2}+e_{m3}+e_{m4}}{4}
\end{equation}
where $e_{mi}$ is the error for the $i^{th}$ model and we use four models namely, logistic regression, random forest, decision tree, and multi layer perceptron as given in~\cite{brenninkmeijer2019synthetic}. The models are trained to learn the tasks of classification on target columns and  regression on target columns with random forests, lasso regression, ridge regression, and ElasticNet.\par

All the 5 metrics discussed above evaluate different aspects of similarity between the real and synthetic data. We further combine all these metrics and propose a unified metric TabSynDex. The proposed TabSynDex score is calculated as given in Eq.~\ref{eq11}.
\begin{equation}
\label{eq11}
    TabSynDex = \frac{S_{basic}+S_{corrl}+S_{pMSE}+S_{cr}+S_{ml}}{5}
\end{equation}

\subsection{Comparison with the Existing Metrics}
The aim of a synthetic tabular data evaluation metric is to produce a similarity value representing the closeness of synthetic data to the real data. In this regard,~\cite{brenninkmeijer2019synthetic} combines several statistical and machine learning metrics. Apart from the resulting scalar value, some additional metrics are reported to highlight other characteristics of the data. It combines all the component scores to get a single metric which can be used for bench-marking the GAN architectures for data synthesis. But the methodology used to calculate this score has few shortcomings: 
\begin{enumerate}
    \item Correlation is used at multiple places to bound the scores between $0$ to $1$. For example, the correlation between the vectors containing the statistics of the real data and synthetic data is computed as part of the metric. We opine that nearness to the~\textit{original data} points should have been accounted for in this metric.
    
    \item Similarly, in the machine learning based metric, the nearness of the synthetic data to the real data is not considered. The metric first mirrors the column association between the real-synthetic data, and calculates the correlation between them. This is a flawed strategy as any two random vectors would yield high correlation if they are sorted.
\end{enumerate}

Table~\ref{tab:metric_aspect_comparison} shows the various desirable attributes (1st row) that an ideal metric should consider. For example, inter-column correlation, distribution comparison, rare class coverage, implicit bound, and suitability to machine learning applications are some desired attributes. We compare the existing metrics in the literature (1st column) with the proposed TabSynDex. It is clear that the proposed TabSynDex is the only metric that covers all the listed attributes in~Table~\ref{tab:metric_aspect_comparison}.

\textbf{Merits of the Proposed TabSynDex:}
The proposed TabSynDex metric has several merits that makes it a better choice for evaluation.

\begin{enumerate}
    \item It can be measured on an absolute scale. Unlike the metrics like RMSE scores of machine learning models, likelihood fitness, Wasserstein Distance, pMSE etc. which can only be compared for two models. The TabSynDex is restricted to the range [0, 1] unlike the metrics mentioned above which restrict their usage for judging a model alone.
    \item It gives a single score cumulative metric encapsulating the various desired qualities in synthetic data. This helps in bench-marking the results across different generative models. Currently, a variety of metrics are used selectively which makes it difficult to compare the existing methods for data synthesis. 
    \item The proposed metric allows fast calculation after every epoch, allowing us to gain useful insights during training. For example, in~CTGAN~\cite{xu2019modeling} the authors train each model for 300 epochs. However, our study shows that CTGAN does not learn at all during training. The model gives similar results to that of the Gaussian Mixture Models which was used to pre-process the data. No training and 100 epochs training give almost identical results. A metric like the proposed TabSynDex would make it possible to notice such behaviour during training. We have shown this in our experiments in Section~\ref{sec:experiments}.
    
    \item The analysis of each component metric of TabSynDex gives insights into what is happening with the model and what might happen. For example, in Section~\ref{sec:experiments}, a decreasing histogram comparison score can hint at an upcoming mode collapse. Similarly, a straight line histogram comparison along with 0 correlation score denotes a mode collapse.
\end{enumerate}

\begin{table}[]
    \centering
    \resizebox{\columnwidth}{!}{
    \begin{tabular}{c|c|c|c|c|c|c}
    \hline
        {} & {Inter-column} & {Distribution} & {Rare class} & {Suitability to} & \multirow{2}{*}{Bounded} & \multirow{2}{*}{Absolute}\\
        {} & {Correlation} & {Comparison} & {Coverage} & {ML Applications} & {} & {}\\
        \hline
        {KL/JS-Divergence~\cite{support_coverage, zhao2022ctab}} & {$\times$} & {\checkmark} & {$\times$} & {$\times$} & {$\times$} & {$\times$} \\
        {Correlation Difference~\cite{zhao2022ctab}} & {\checkmark} & {$\times$} & {$\times$} & {$\times$} & {\checkmark} & {\checkmark} \\
        {Wasserstein Distance~\cite{zhao2022ctab}} & {$\times$} & {\checkmark} & {$\times$} & {$\times$} & {$\times$} & {$\times$} \\
        {ML Performance~\cite{xu2018synthesizing, xu2019modeling}} & {$\times$} & {$\times$} & {$\times$} & {\checkmark} & {$\times$} & {$\times$} \\
        {pMSE~\cite{pmse_expected}} & {$\times$} & {\checkmark} & {$\times$} & {$\times$} & {$\times$} & {$\times$} \\
        {TabSynDex} & {\checkmark} & {\checkmark} & {\checkmark} & {\checkmark} & {\checkmark} & {\checkmark} \\
        \hline
    \end{tabular}
    }
    \caption{A comparison between the proposed TabSynDex and the existing metrics for synthetic tabular data analysis. The desirable attributes for each metric is listed in the first row.}
    \label{tab:metric_aspect_comparison}
\end{table}

\subsection{Models for Tabular Data Generation}
\label{sec:models}
We use several generative adversarial networks for tabular data synthesis. We conduct experiments on~TVAE~\cite{kingmaauto,xu2019modeling}, CTGAN~\cite{xu2019modeling} and TGAN~\cite{xu2018synthesizing} with Gaussian Mixture Model (GMM) pre-processing. We also execute experiments on WGAN-GP~\cite{gulrajaniWGANGP2017,arjovskyWGAN2017} in order to observe the results without GMM pre-proccesing. Furthermore, we introduce certain variations in the original TGAN and train the models to see if any minute change in the model makes any difference to the otherwise not so encouraging training graphs and performance. The following are the variations of TGAN that we implement in this paper.\par
\textbf{TGAN with GRU:} We replaced the LSTM with GRU~\cite{chung2014empirical} in the generator. Rest of the specifics remain same to that of TGAN.\par
\textbf{TGAN with a bidirectional generator:} In the original TGAN, each column receives the context of the earlier columns for generation. We introduce a bidirectional generator~\cite{donahue2016adversarial} in which each output is generated after combining the respective column features (dense layers) from both the forward pass and backward pass models.\par
\textbf{TGAN with Variational GMM (VGMM) preprocessing:} We replace GMM with VGMM as given in~\cite{xu2019modeling}.\par
\textbf{TGAN with bidirectional generator and VGMM preprocessing:} We replace the GMM with VGMM in the TGAN model with a bidirectional generator.\par

\textit{It is to be noted that t\textbf{he detailed analysis of these four variants of tabular GANs and corresponding results also serve as a natural ablation study of our work.}}

\section{Datasets and Experiment Setting}
We analyse the performance of TGAN, CTGAN, WGAN-GP, TVAE, and several modifications of TGAN on datsets of various sizes and nature. We then evaluate their performance in various scenarios with five constituent metrics of the proposed TabSynDex for tabular data synthesis.

\subsection{Datasets}
We compare the performance and training progress of all the models on the following datasets: (i) Concrete Compressive Strength Data~\cite{concrete_data} - 1,030 rows, 9 columns. (ii) Wine Quality Data~\cite{wine_data} - 4,898 rows, 13 columns (We use only the white wine data). (iii) Online News Popularity Data Set~\cite{news_data} - 39,797 rows, 61 columns. (iv) Combined Cycle Power Plant Data Set~\cite{electrical_data} - 9,568 rows, 4 columns.\par

We have used a variety of datasets with various number of data points and columns. The variety in these datasets help us analyze the results in different scenarios such as (i) which model performs well with extremely less amounts of data, (ii) which model performs best when a decent number of samples are available, (iii) does the number of samples required for decent performance depend on the number of columns? (as with increasing number of columns, higher number of relations need to be covered by the model).

\begin{table}[t]
\centering
\resizebox{\columnwidth}{!}{
\begin{tabular}{c|c|c|c|c|c}
    \hline
    \multirow{3}{*}{Data}& \multirow{3}{*}{Number of rows}& \multicolumn{4}{c}{TabSynDex Score}\\ \cline{3-6}
         &  & \multicolumn{4}{c}{\% of real data treated as synthetic}\\ \cline{3-6}
        & &10\%&25\%&50\%&100\%\\ \hline
        Concrete~\cite{concrete_data} &1030&0.768&0.869&0.914&0.894\\ \hline
        News Popularity~\cite{news_data} &39644&0.891&0.916&0.901&0.898\\ \hline
        Wine Quality~\cite{wine_data} &4898&0.867&0.911&0.925&0.938\\ \hline
        Power Plant~\cite{electrical_data} &9568&0.946&0.961&0.981&0.969\\ \hline
    \end{tabular}
}
\caption{Experiment for sanity check of the TabSynDex metric for tabular data synthesis evaluation. The real dataset is divided into different subsets to check the similarity between them using TabSynDex.~\textit{A higher TabSynDex score is better}}
\label{tab:subset_similarity}
\end{table}

\subsection{Sanity check of the metric for tabular data synthetic}
\label{sec_sanity_check}
We first perform a~\textit{sanity check} on the evaluation metric by observing its behaviour on two different subsets of a real dataset. The metric should ideally give a high similarity score between these two subsets. We shuffle and partition the original dataset into two equal parts, using one part as the real dataset, and treating the other as the synthesized dataset. We sample different proportions of data from the synthetic dataset to observe the effect of \textit{sample size} on the overall score. We hypothesize that a small set of samples will show less similarity with the real data as it covers a small variety of samples. Besides, it will have high degree of randomness which would lead to less score on all components of the metric.

\subsection{Experiment Settings}
We use the GAN and VAE-based models as mentioned in Section~\ref{sec:models} to generate synthetic data. We evaluate their training progression to observe the evolution in the quality of data generated over the number of epochs. We also notice the highest similarity score obtained during training. The obtained similarity score is compared with the score obtained by a Bayesian based data synthesis method DataSynthesizer~\cite{DataSynthesizer}.\par

For CTGAN and TGAN, the default hyper parameters are adopted. These hyper parameters are maintained across different modifications of these models throughout our experiments. In WGAN-GP, the min-max scaling is used for non categorical columns and one hot encoding is used for categorical columns.~For TVAE, we follow the hyper parameters as in ~\cite{xu2019modeling}. Almost all the models either show no sign of improvement or converge by 50 epochs on each dataset. Therefore, we present the results for 50 epochs. For DataSynthesizer, we use the~\textit{correlated\_attribute} mode and keep the degree of the Bayesian network as 2.

\begin{figure}[]
\centering
\begin{minipage}{\linewidth}
  \centering
  \includegraphics[width=0.9\linewidth]{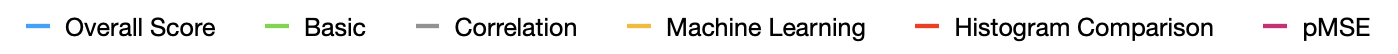}
  \begin{tabular}{cc}
  \includegraphics[width=.45\linewidth]{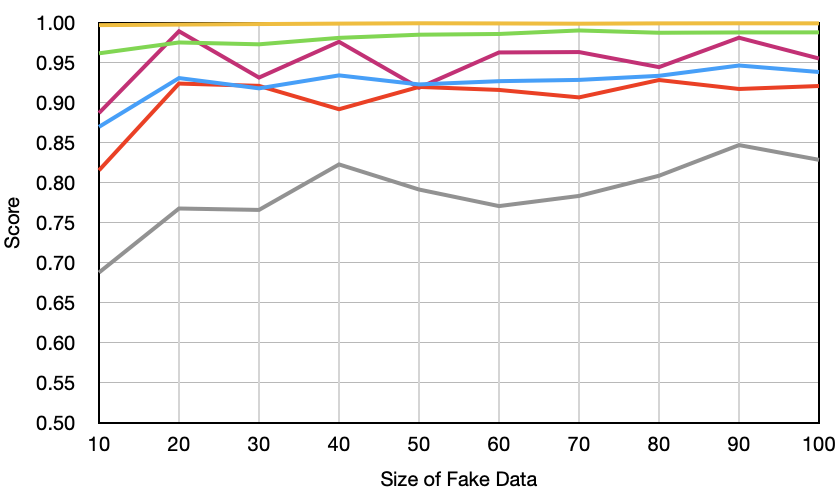}
    & \includegraphics[width=.45\linewidth]{img/Data_Scores/wine1.png}\\
      \scriptsize{i. Concrete Data} & \scriptsize{ii. Wine Quality Data}\\
    \includegraphics[width=.45\linewidth]{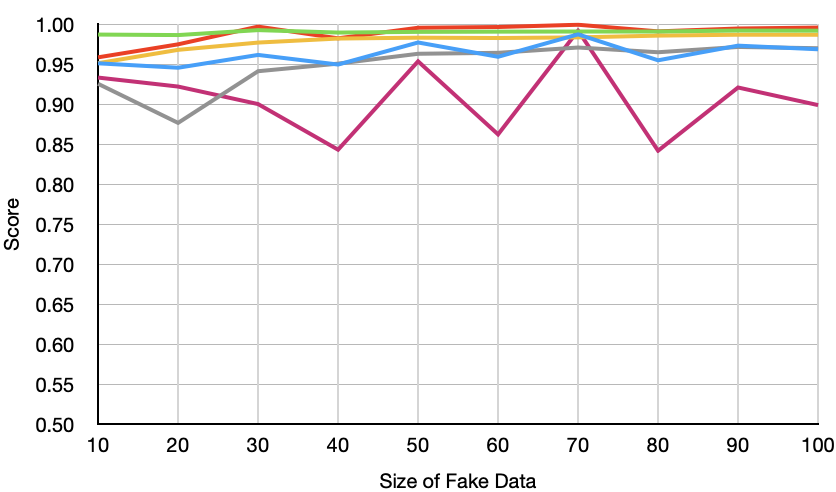}
    & \includegraphics[width=.45\linewidth]{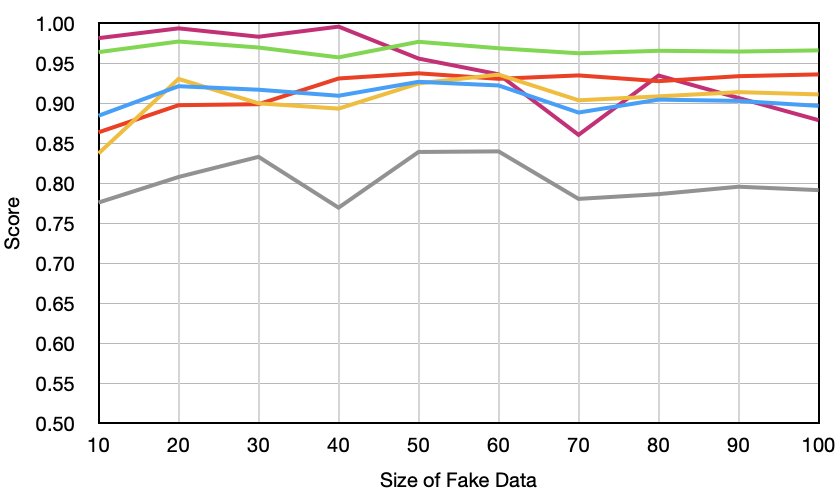}\\
 \scriptsize{iii. Electrical Power Plant Data} & \scriptsize{iv. News Data} \\ 
  \end{tabular}
  \end{minipage}
 \caption{Cross-validation analysis with the proposed TabSynDex metric. The real dataset is divided into two sets. The synthetic data is generated by using one set of data. The generated synthetic data is then compared with the other set of real data.}
 \vspace{-1em}
\label{fig:cross-val}
\end{figure}

\begin{figure*}
 \begin{minipage}{\linewidth}
  \centering
   \includegraphics[width=0.9\linewidth]{legends.png}
  \begin{tabular}{cccc}
  \includegraphics[width=.23\linewidth]{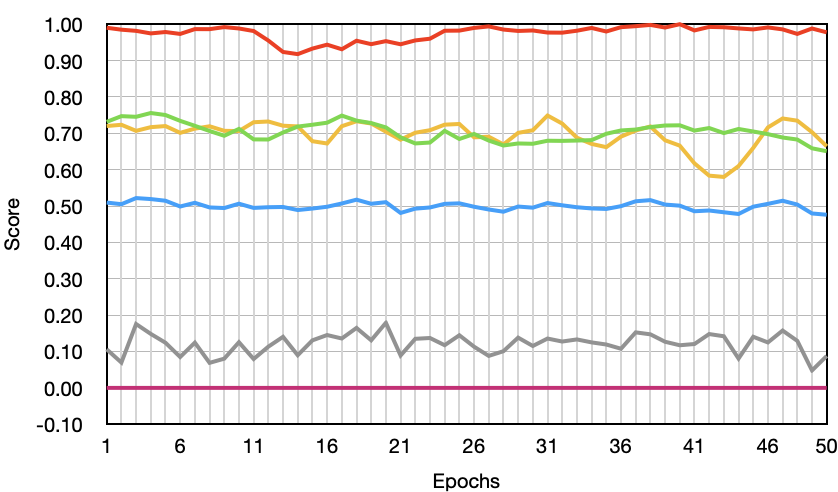}
    & \includegraphics[width=.23\linewidth]{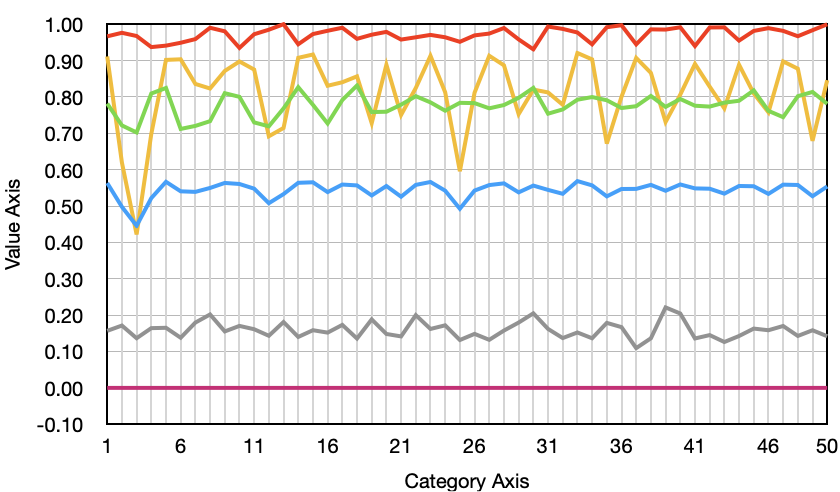}
    & \includegraphics[width=.23\linewidth]{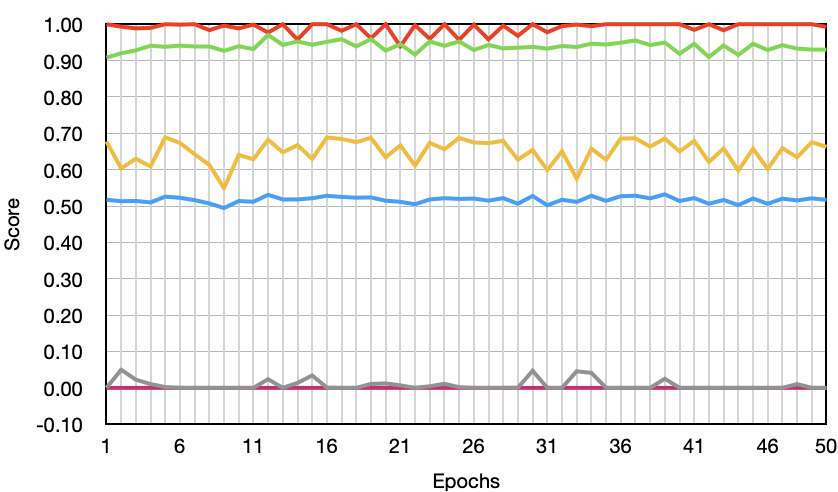}
    & \includegraphics[width=.23\linewidth]{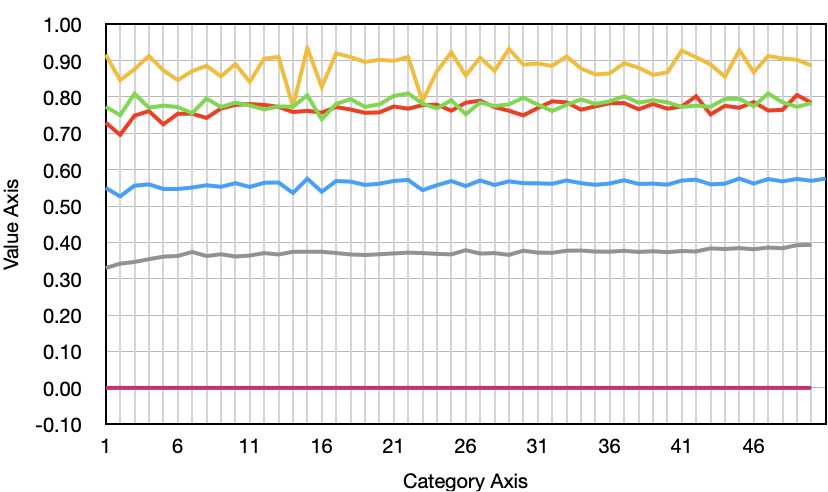} \\
  \scriptsize{i. Concrete Data} & \scriptsize{ii. Wine Quality Data} & \scriptsize{iii. Electrical Power Plant Data} & \scriptsize{iv. News Data} \\  
  \end{tabular}
  \subcaption{CTGAN}
  \vspace{-1em}
  \label{fig:real-a}
  \end{minipage}\par\bigskip
  \begin{minipage}{\linewidth}
  \centering
  \begin{tabular}{cccc}
  \includegraphics[width=.23\linewidth]{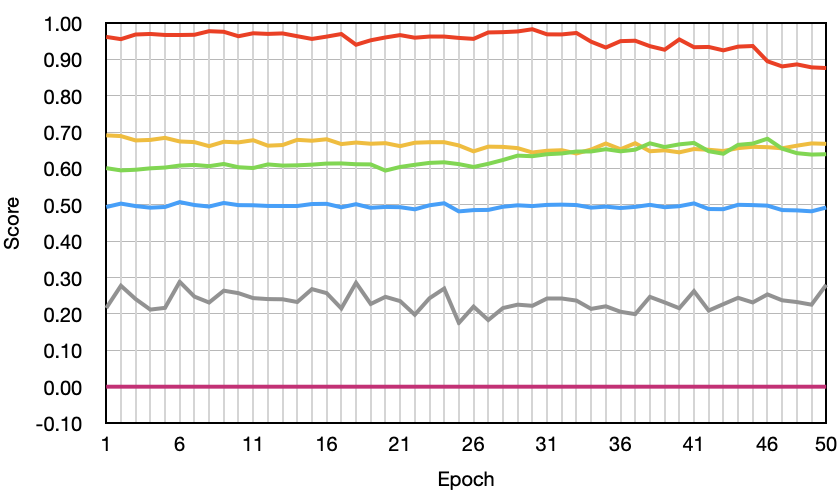}
    & \includegraphics[width=.23\linewidth]{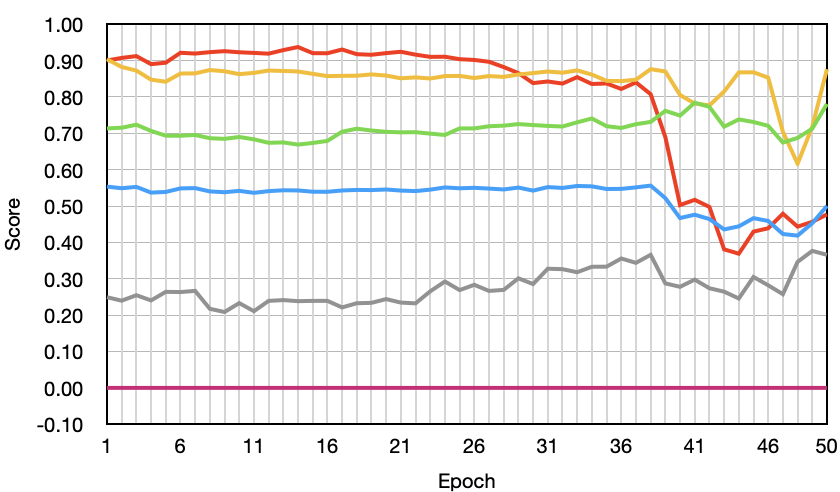}
    & \includegraphics[width=.23\linewidth]{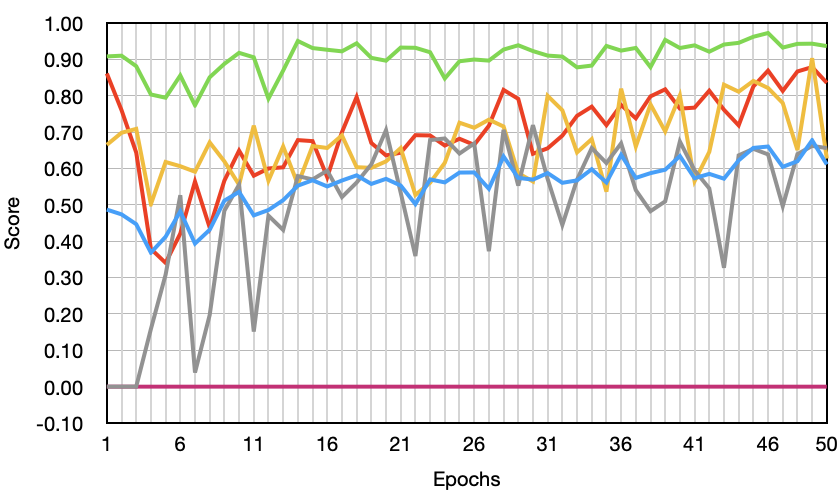}
    & \includegraphics[width=.23\linewidth]{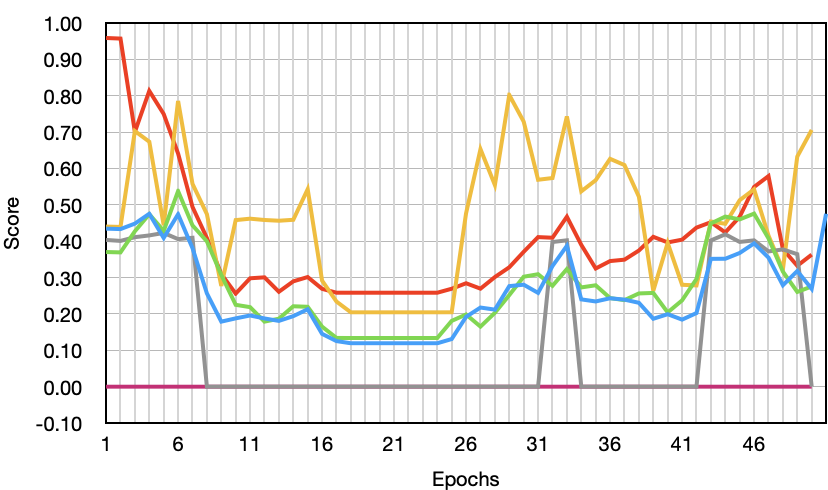} \\
  \scriptsize{i. Concrete Data} & \scriptsize{ii. Wine Quality Data} & \scriptsize{iii. Electrical Power Plant Data} & \scriptsize{iv. News Data} \\  
  \end{tabular}
  \subcaption{TGAN}
  \vspace{-1em}
  \label{fig:real-b}
  \end{minipage}\par\bigskip
  \begin{minipage}{\linewidth}
  \centering
  \begin{tabular}{cccc}
  \includegraphics[width=.23\linewidth]{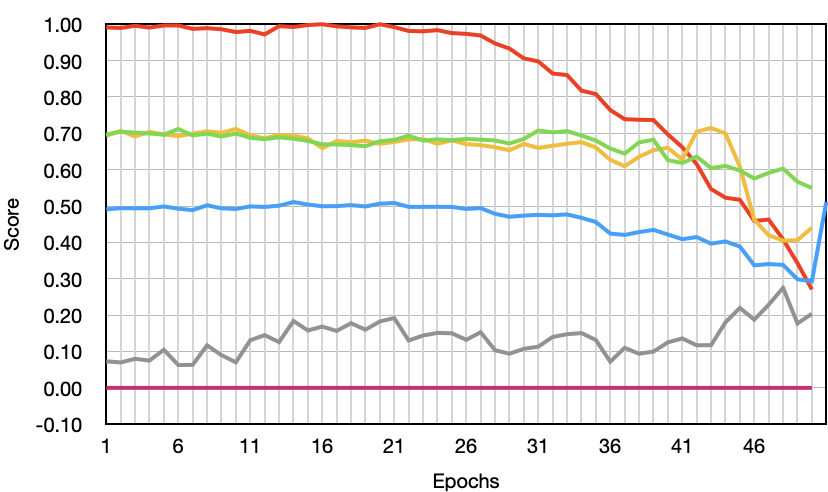}
    & \includegraphics[width=.23\linewidth]{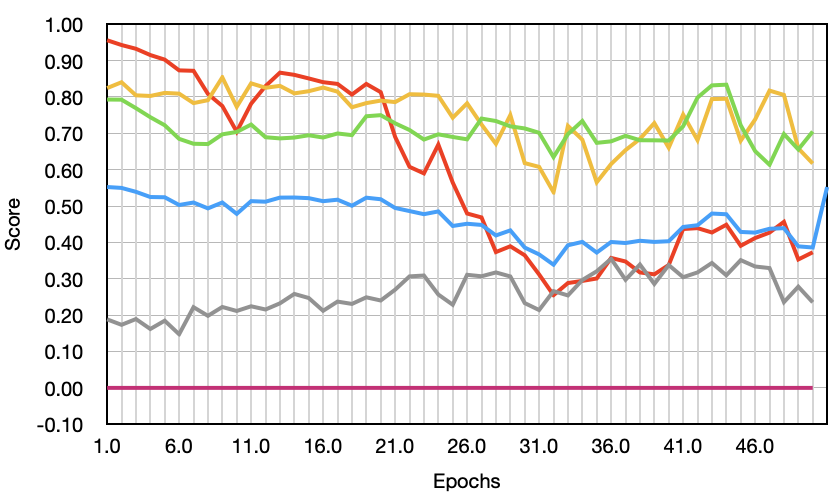}
    & \includegraphics[width=.23\linewidth]{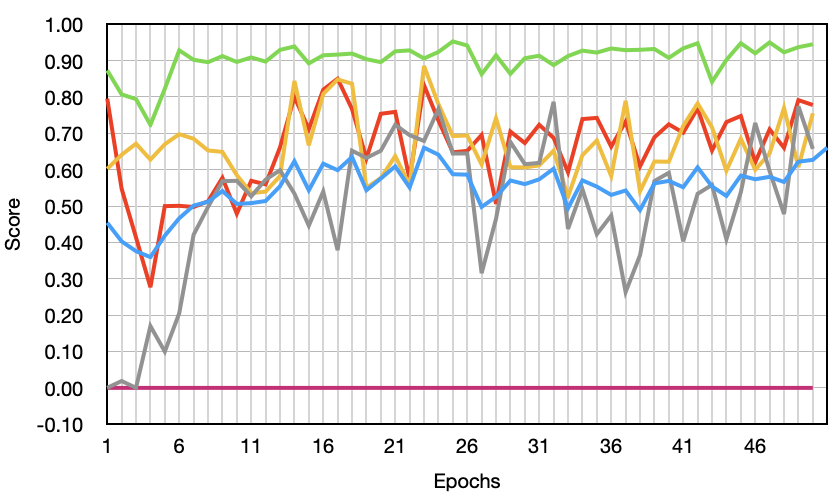}
    & \includegraphics[width=.23\linewidth]{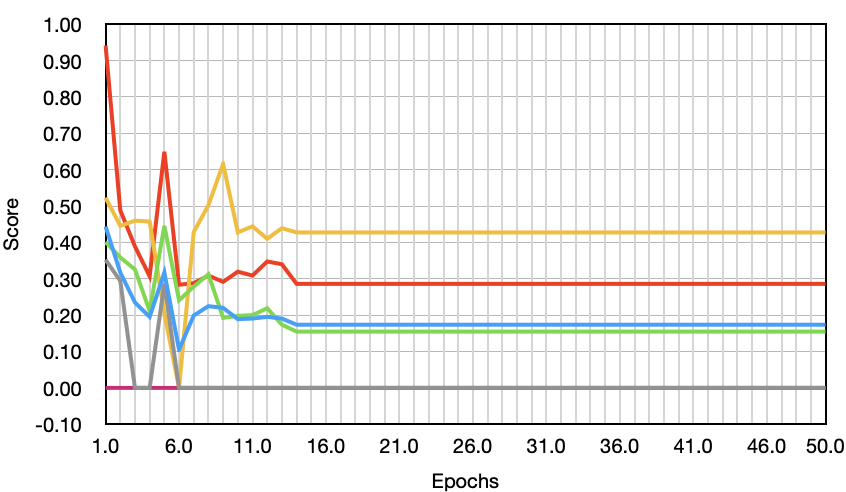} \\
  \scriptsize{i. Concrete Data} & \scriptsize{ii. Wine Quality Data} & \scriptsize{iii. Electrical Power Plant Data} & \scriptsize{iv. News Data} \\  
  \end{tabular}
  \subcaption{TGAN with GRU}
  \vspace{-1em}
  \label{fig:real-c}
  \end{minipage}
\caption{TabSynDex metric comparison for the synthetic data generated by CTGAN, TGAN, and TGAN with GRU. We observe that none of the methods generate 'good quality' synthetic data i.e., having distribution close to the real data.}
\label{fig:real}
\end{figure*}

\begin{figure*}
\centering
\begin{minipage}{\linewidth}
  \centering
  \includegraphics[width=0.7\linewidth, height=0.5cm]{legends.png}
  \begin{tabular}{cccc}
  \includegraphics[width=.23\linewidth]{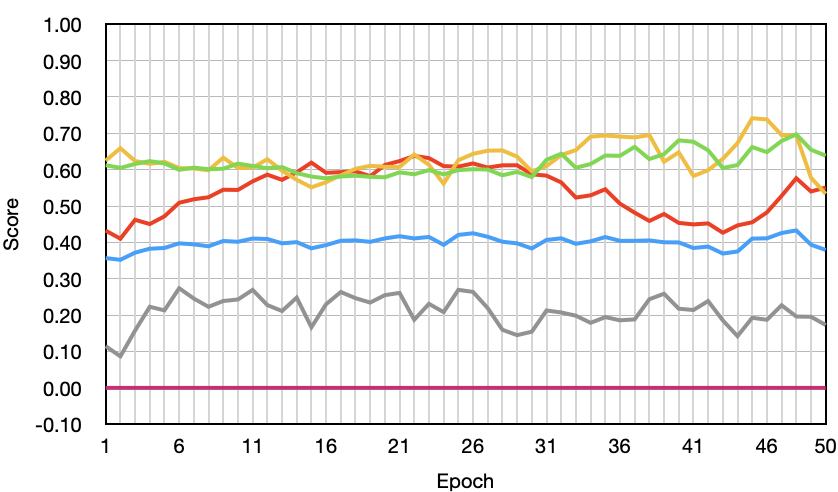}
    & \includegraphics[width=.23\linewidth]{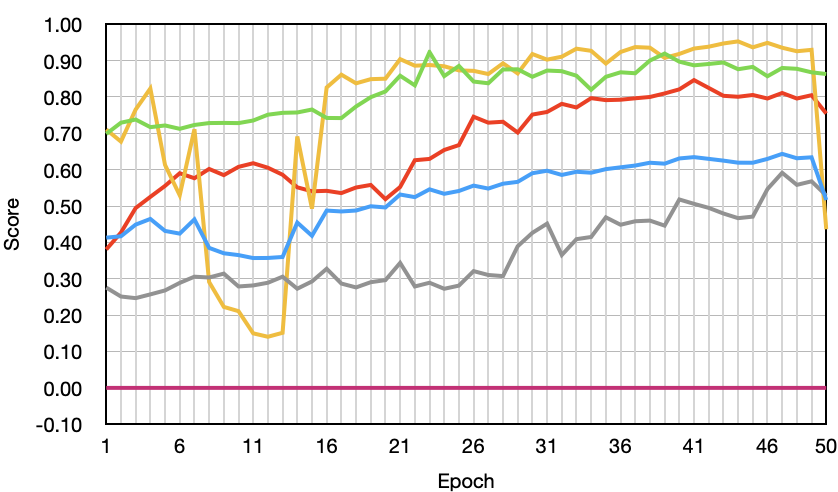}
    & \includegraphics[width=.23\linewidth]{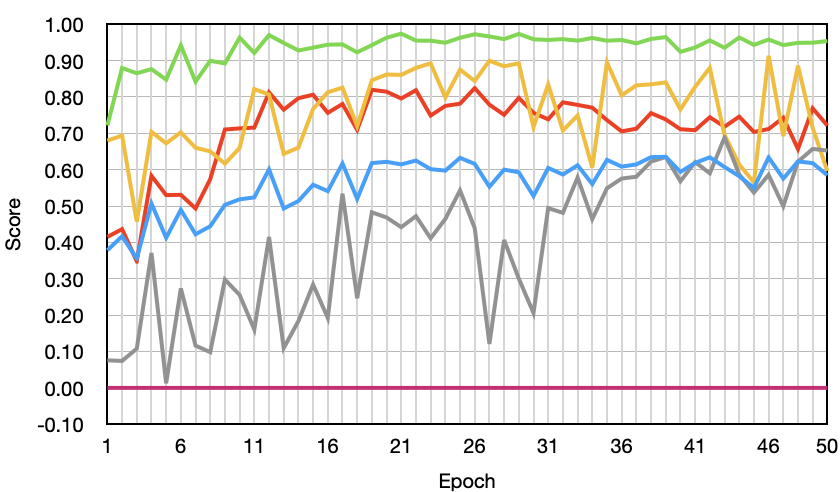}
    & \includegraphics[width=.23\linewidth]{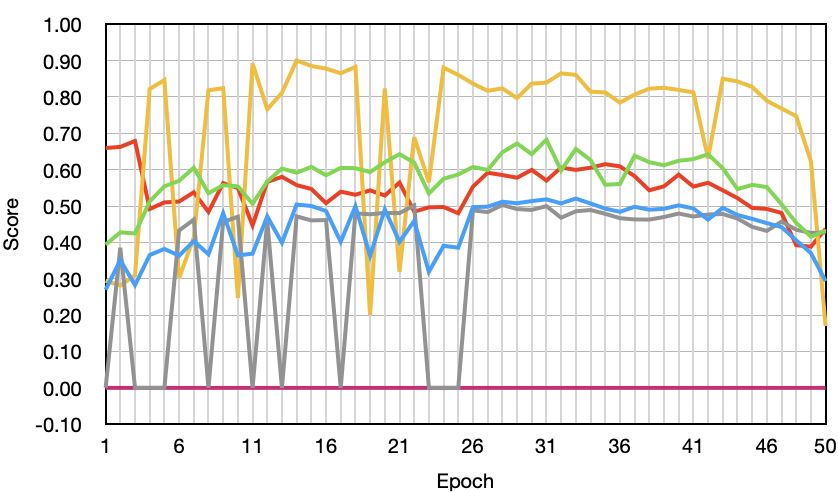}\\
  \scriptsize{i. Concrete Data} & \scriptsize{ii. Wine Quality Data} & \scriptsize{iii. Electrical Power Plant Data} & \scriptsize{iv. News Data} \\  
  \end{tabular}
  \subcaption{Bidirectional TGAN}
  \vspace{-1em}
  \label{fig:synthetic-a}
  \end{minipage}\par\bigskip
  \begin{minipage}{\linewidth}
  \centering
  \begin{tabular}{cccc}
  \includegraphics[width=.23\linewidth]{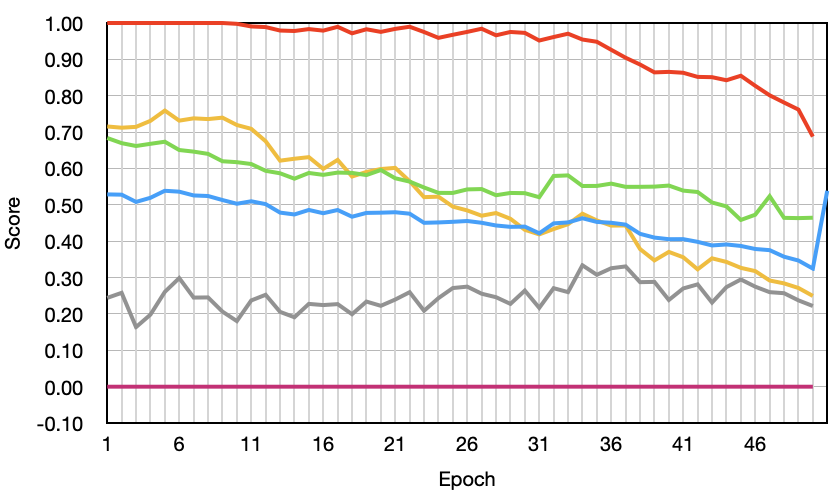}
    & \includegraphics[width=.23\linewidth]{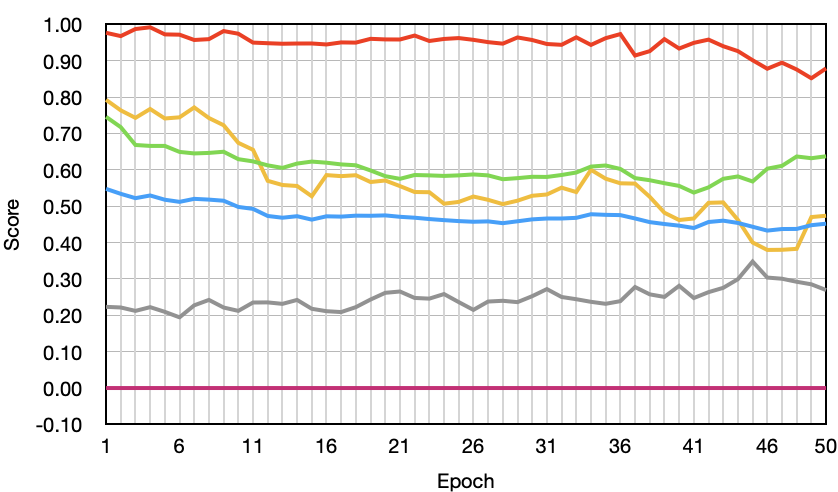}
    & \includegraphics[width=.23\linewidth]{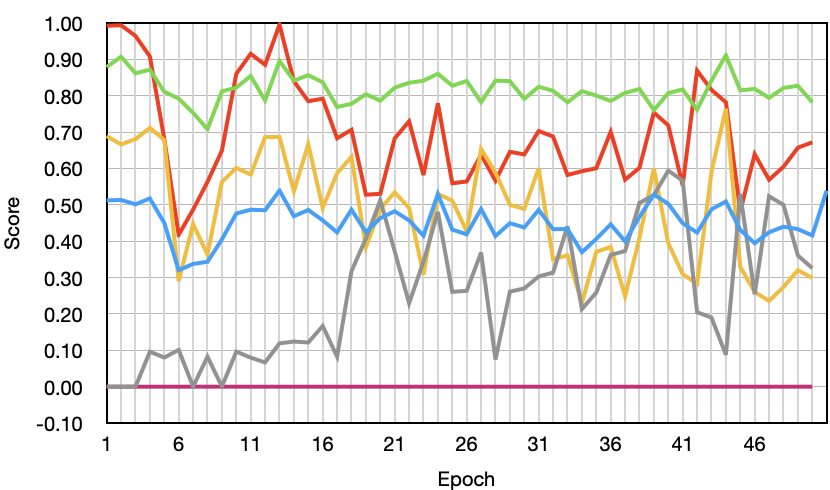}
    & \includegraphics[width=.23\linewidth]{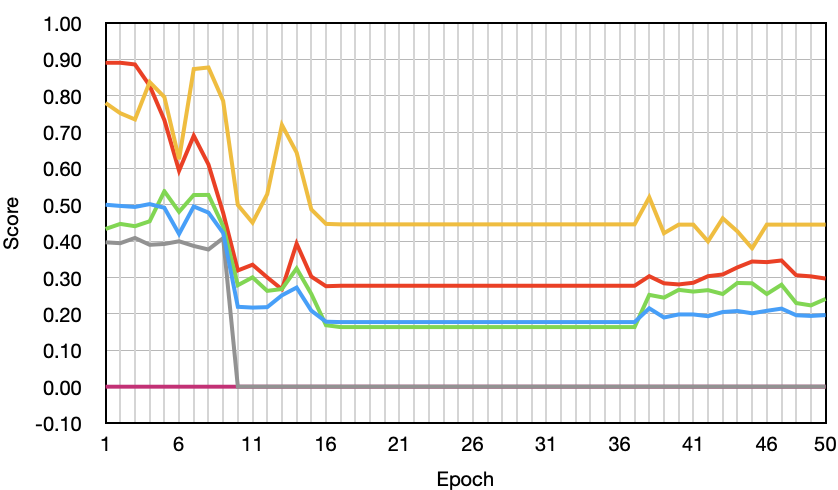} \\
  \scriptsize{i. Concrete Data} & \scriptsize{ii. Wine Quality Data} & \scriptsize{iii. Electrical Power Plant Data} & \scriptsize{iv. News Data} \\  
  \end{tabular}
  \subcaption{TGAN with VGMM Preprocessing}
  \vspace{-1em}
  \label{fig:synthetic-b}
  \end{minipage}\par\bigskip
  \begin{minipage}{\linewidth}
  \centering
  \begin{tabular}{cccc}
  \includegraphics[width=.23\linewidth]{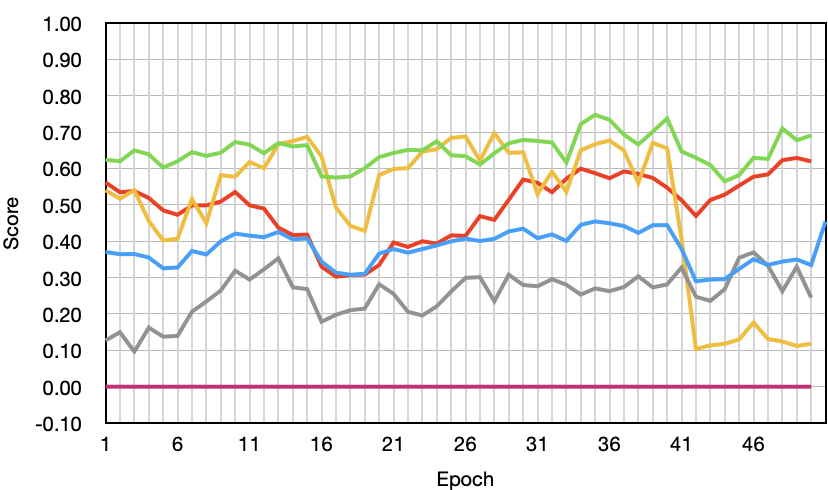}
    & \includegraphics[width=.23\linewidth]{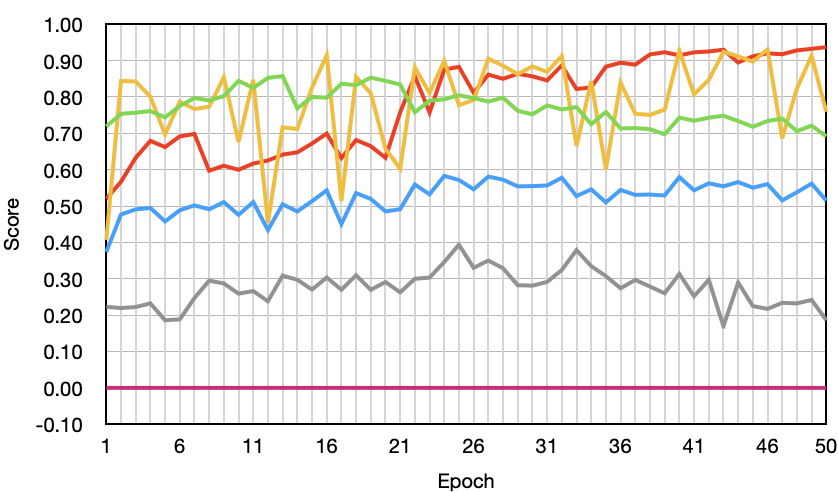}
    & \includegraphics[width=.23\linewidth]{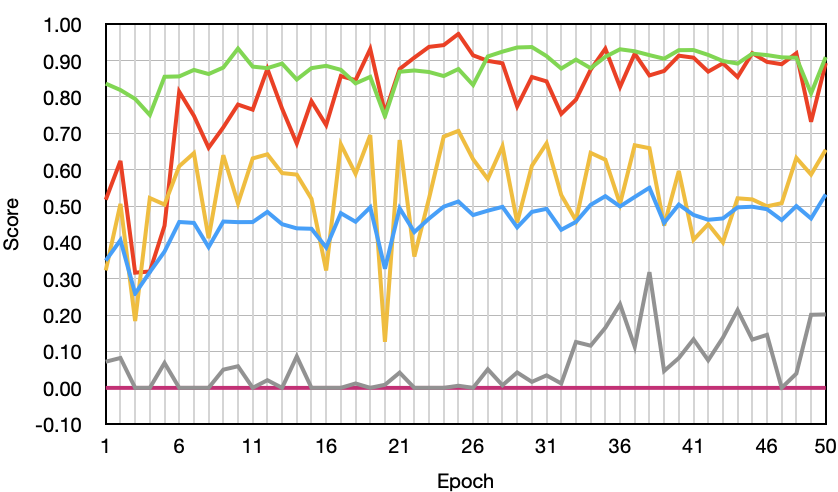}
    & \includegraphics[width=.23\linewidth]{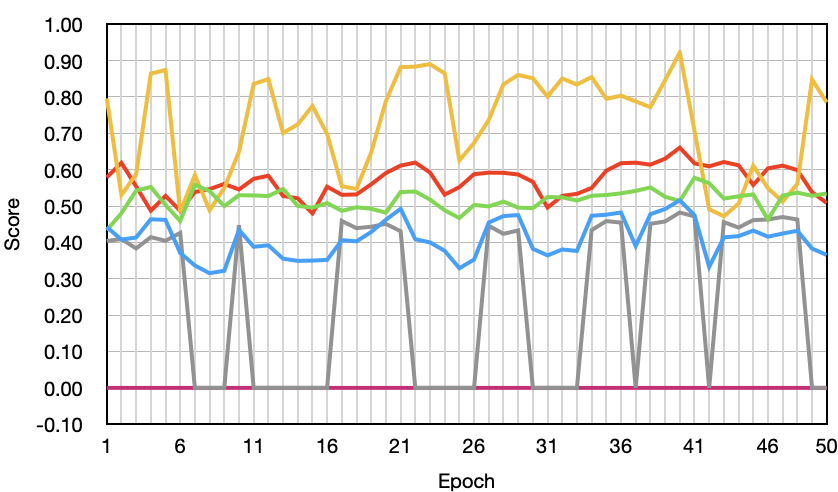} \\
  \scriptsize{i. Concrete Data} & \scriptsize{ii. Wine Quality Data} & \scriptsize{iii. Electrical Power Plant Data} & \scriptsize{iv. News Data} \\  
  \end{tabular}
  \subcaption{Bidirectional TGAN with VGMM Preprocessing}
  \vspace{-1em}
  \label{fig:synthetic-c}
  \end{minipage}\par\bigskip
  \begin{minipage}{\linewidth}
  \centering
  \begin{tabular}{cccc}
  \includegraphics[width=.23\linewidth]{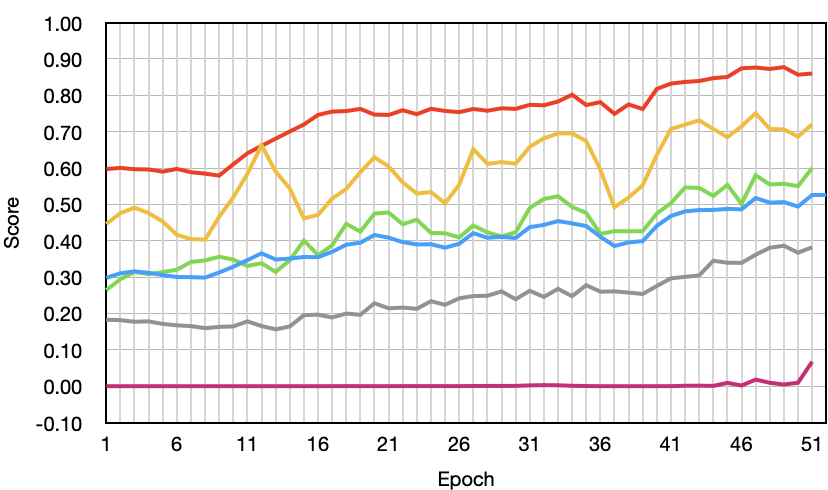}
    & \includegraphics[width=.23\linewidth]{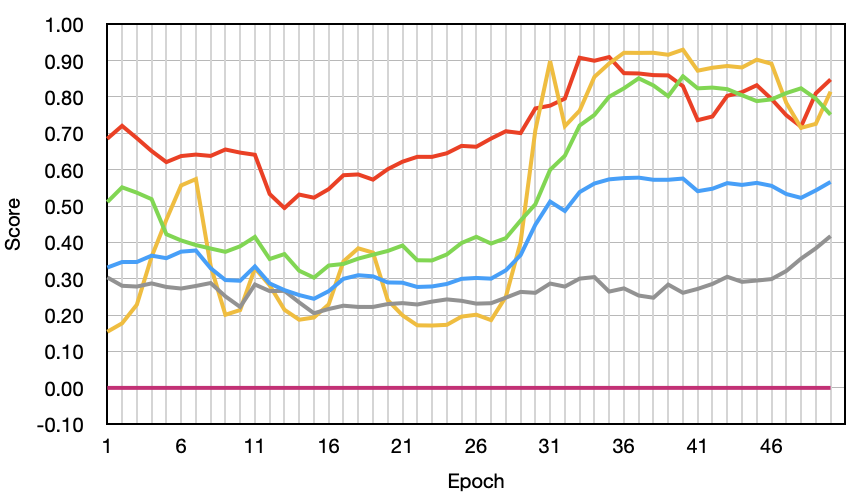}
    & \includegraphics[width=.23\linewidth]{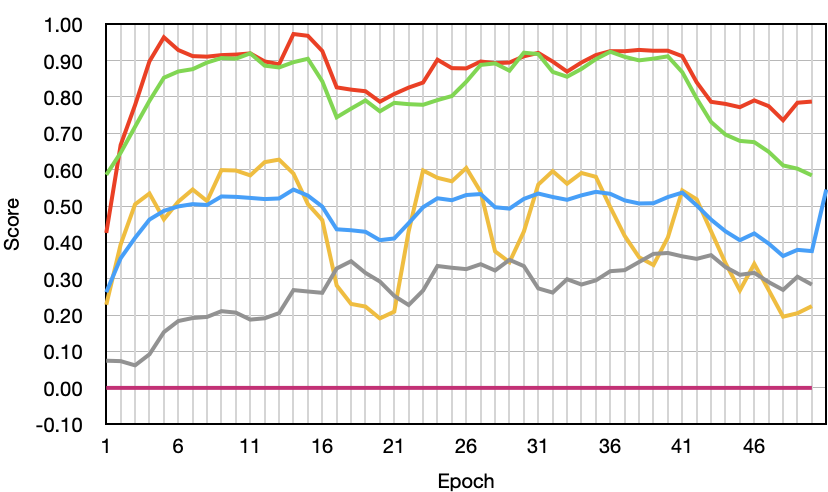}
    & \includegraphics[width=.23\linewidth]{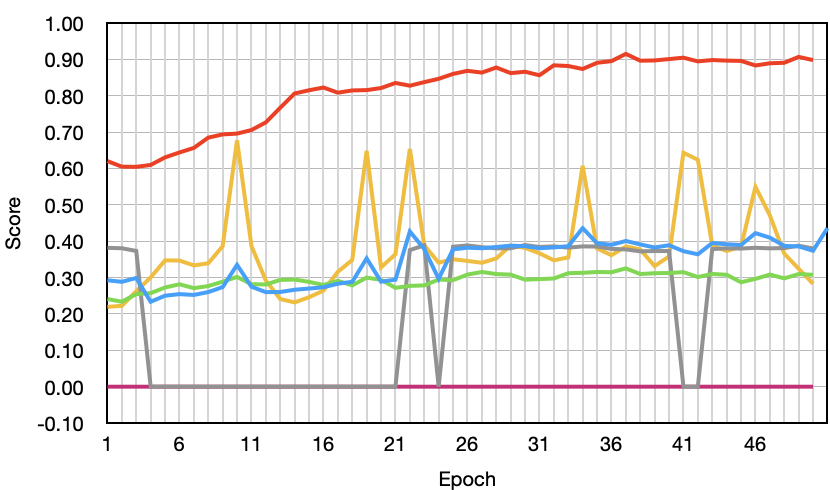}\\
  \scriptsize{i. Concrete Data} & \scriptsize{ii. Wine Quality Data} & \scriptsize{iii. Electrical Power Plant Data} & \scriptsize{iv. News Data} \\  
  \end{tabular}
  \subcaption{WGAN-GP}
  \vspace{-1em}
  \label{fig:synthetic-d}
  \end{minipage}
\caption{TabSynDex score of synthetic tabular data generated by Bidirectional TGAN, TGAN with VGMM Preprocessing, Bidirectional TGAN with VGMM Preprocessing, and WGAN-GP. The quality of the synthesized data is not very good in most of the methods. Moreover, there is negligible learning over the epochs in almost every case except WGAN-GP.}
\label{fig:synthetic}
\end{figure*}

\begin{table}[t]
    \centering
    \resizebox{\columnwidth}{!}{
\scalebox{0.85}{\begin{tabular}{c|c|c|c|c}
    \hline
        \multicolumn{5}{c}{TabSynDex Score} \\
         \hline
         Method & Concrete & Wine Quality & Power Plant & News\\
         \hline
         DataSynthesizer~\cite{DataSynthesizer} & 0.47 &  0.49   & 0.52 &  0.38\\
         TVAE~\cite{kingmaauto,xu2019modeling} & 0.72 & 0.45 & 0.65 & 0.40 \\
         CTGAN~\cite{xu2019modeling}   &0.52&  0.57   & 0.53&  0.58\\
         TGAN (LSTM)~\cite{xu2018synthesizing}& 0.51 & 0.56  &  0.68 & 0.48\\
         TGAN (GRU)~\cite{chung2014empirical} &0.51 &0.55 & 0.66 & 0.44\\
         TGAN (Bi)~\cite{xu2019modeling,donahue2016adversarial}    &0.43 &0.64 & 0.64 &0.52\\
         TGAN (VGMM)~\cite{xu2019modeling} & 0.54 & 0.55 & 0.54 & 0.50\\
         TGAN (Bi, VGMM)~\cite{xu2019modeling} & 0.45 & 0.58 & 0.55 & 0.52\\
         WGAN-GP~\cite{gulrajaniWGANGP2017,arjovskyWGAN2017} & 0.53 & 0.58 & 0.55 & 0.44\\
         \hline
    \end{tabular}}
    }
    \caption{Comparative analysis of the tabular synthetic data generated by different models. The TabSynDex score on the test set is reported on 4 datasets.}
    \label{tab:results}
\vspace{-1em}
\end{table}

\section{Experimental Results and Analysis}
\label{sec:experiments}
\subsection{Cross-validation Analysis}
\label{sec_cross_val}
In the first experiment, we perform a sanity check on the proposed TabSynDex metric. The idea is that if two non-overlapping subsets of a real dataset are compared then the TabSynDex metric should be close to 1 (the highest score). Therefore, we take different proportions of the real dataset and compute the TabSynDex between them to check the correctness of the proposed metric. We create two subsets of the real dataset: \textit{subset1} and \textit{subset2}. The TabSynDex score between these subsets will be computed. Table~\ref{tab:subset_similarity} shows the overall TabSynDex score when $10\%$, $25\%$, $50\%$, and $100\%$ of samples are used from \textit{subset2}. We sample various proportions of data from \textit{subset2} to evaluate the effect of data size on the similarity metric. We observe that using a small set of data leads to poor TabSynDex score. Ideally, all the different proportion of samples ($10\%$, $25\%$, $50\%$, and $100\%$) should give approximately the same similarity score as they all belong to the real dataset. However, in practice, using a small proportion of data leads to lower similarity score due to their inherent randomness. Thus, the amount of data present in \textit{subset2} affects the similarity metric. Having more number of data points will reduce the randomness and lead to a better understanding about the characteristics of the data. In Fig.~\ref{fig:cross-val}, the TabSynDex and its constituent sub-metrics are plotted for: concrete data, wine quality data, electrical power plant data, and news data. This further gives a more comprehensive view of how the number of samples effect the similarity score and its components.\par

\subsection{Quantitative Analysis with different Generative Methods}
A good synthetic data generator should produce data that is similar to the data present in \textit{subset2} as discussed in Section~\ref{sec_cross_val}. The~\textit{subset1} is used by the model to learn the distribution of the data. After fine-tuning the model, synthetic data is generated. We now use \textit{subset2} to compare with the synthetic data. We show the results (on all four datasets) of the CTGAN~\cite{xu2019modeling}, TGAN~\cite{xu2018synthesizing}, and TGAN with GRU~\cite{chung2014empirical} in Fig.~\ref{fig:real}. The results of bidirectional TGAN, TGAN with VGMM preprocessing~\cite{xu2019modeling}, bidirectional TGAN with VGMM preprocessing~\cite{xu2019modeling}, and WGAN-GP~\cite{gulrajaniWGANGP2017,arjovskyWGAN2017} is depicted in Fig.~\ref{fig:synthetic}.

\subsubsection{\textbf{CTGAN}} Fig.~\ref{fig:real-a} shows the results of CTGAN~\cite{xu2019modeling} on four datasets. The datasets having less number of samples (i.e., \textit{concrete data}, and \textit{wine quality data}) cause more variation. The comparison between the small test subsets lead to lower similarity score with lot of fluctuations at different epochs. The CTGAN trained of the larger datasets (i.e., \textit{power plant data} and \textit{news data}) give around $90\%$ similarity. All the component scores are quite good, hinting that good quality synthetic data will score well as the distribution is very similar to real data. Furthermore, all the component scores of TabSynDex remain almost constant with respect to the increasing epochs. This leads to a constant overall score on all the 4 datasets. We note that the $S_{pMSE}$ score stays constant at 0 and never changes. In fact $S_{pMSE}$ never rises above 0 for any model or dataset (except for 1 case which we will discuss later). This means the data synthesized by all the models can easily be differentiated from real data with a linear regression model. Essentially, all the existing synthetic data generation methods fail to produce data that is non-differentiable from real data. Our study shows that synthesizing completely non-differentiable real data remains an unsolved problem and a challenging future research work.

\subsubsection{\textbf{TGAN}} The results using TGAN~\cite{xu2018synthesizing} method is shown in Fig.~\ref{fig:real-b}. On \textit{concrete data}, the results are similar to CTGAN i.e., the scores are constant and training does not have its intended effect of improving the quality of data generated by the model. On \textit{wine quality data}, the scores are constant till the $37^{th}$ epoch after which there is sharp decline in histogram comparison score $S_{cr}$. This could be due to a decrease in the diversity of data produced hinting at a potential mode collapse. We verify the effect of training further with more epochs later in this section. TGAN on \textit{electrical power plant data} is the first instance where training has the intended effect of increasing the quality of generated data. This is visible through all the scores (except $S_{pMSE}$). On \textit{news data}, the correlation score $S_{corr_l}$ goes zero for some intervals of epochs, which is mostly because of mode collapse. 
%($log_{10}(0)$ would be undefined but we cap the score in such cases to 0).

\subsubsection{\textbf{TGAN with GRU}} 
We modify the TGAN by replacing the LSTM cells with GRU~\cite{chung2014empirical} cells. Fig.~\ref{fig:real-c} represents the results for this model. All the component scores decline with respect to TGAN. Specially, the histogram comparison score $S_{cr}$ degrades substantially with increasing epochs. As stated earlier, this may hint at an upcoming mode collapse on \textit{concrete} and \textit{wine quality data}. This is evident from the graph on \textit{news data} where the decrease in all the scores (except $S_{pMSE}$), especially histogram comparison score $S_{cr}$ is followed by a mode collapse on the $6^{th}$ epoch. The model obtains slightly better results on \textit{power plant data} in comparison to the original TGAN.

\subsubsection{\textbf{Bidirectional TGAN}}
We augment another variant of TGAN with bidirectional generator~\cite{donahue2016adversarial}. Fig.~\ref{fig:synthetic-a} shows that a bidirectional TGAN performs better than the original TGAN on \textit{wine quality} and \textit{news data}. In \textit{wine quality data}, the component scores gradually increase till the last of the 50 epochs. All the component scores are quite stable. The performance is even better on \textit{news data}. All the component scores (except $S_{pMSE}$) improve by a good margin in comparison to TGAN. Some of the scores are still unstable with correlation score $S_{corr_l}$ touching zero quite a few times in the first 30 epochs. This means the model might keep collapsing and recover every now and then. The performance is almost same on the \textit{power plant data}, and worse on the \textit{concrete data}. On \textit{concrete data}, the scores are stable, but they are less than those obtained with original TGAN.

\subsubsection{\textbf{TGAN with VGMM Preprocessing}}
We further study the TGAN by replacing the preprocessing method GMM with variational GMM (VGMM)~\cite{xu2019modeling}. Fig.~\ref{fig:synthetic-b} shows the results of TGAN with VGMM preprocessing. In all four datasets, more epochs of training leads to a decrease in the quality of generated data. In the \textit{news dataset}, we observe how the scores decrease and the model collapse at the $10^{th}$ epoch. The results are also worse than that of CTGAN. This shows that adding VGMM had no positive impact on the model. This is contrary to what is shown in~\cite{xu2019modeling}. Xu et al.~\cite{xu2019modeling} claim that introduction of VGMM leads to improved performance with traditional metrics. However, evaluating with the proposed TabSynDex gives opposite conclusion i.e., using VGMM degrades the generated data quality. As already established through sanity check with real data comparison (in Section~\ref{sec_sanity_check} and Section~\ref{sec_cross_val}), the proposed TabSynDex gives a better measurement of synthetic data quality. 

\subsubsection{\textbf{Bidirectional TGAN with VGMM Preprocessing}}
We further combine VGMM preprocessing with bidirectional TGAN~\cite{xu2019modeling}. Fig.~\ref{fig:synthetic-c} shows the obtained results. The results are mixed again. The performance when compared to the original TGAN is worse on the \textit{concrete data}. If the small dip at the end of TGAN is ignored, the histogram comparison score $S_{cr}$ on the \textit{wine quality data} is also lower than TGAN. Unlike in the original TGAN graph, the metrics for the \textit{power plant data} shows no sign of any improvement with increasing epochs. Only for the \textit{news dataset}, this model performs better than the original TGAN. But it is no better than that of bidirectional TGAN. The performances are marginally less in \textit{wine quality} and \textit{power plant data}. This is again contrary to the analysis presented in~\cite{xu2019modeling} which shows VGMM improves the synthetic data quality. The performance of CTGAN is superior and more stable in comparison to this model. 

\subsubsection{\textbf{WGAN-GP}}
We notice that all the previously discussed models with Gaussian mixture model (GMM) or VGMM preprocessing usually begin with a lot of generative knowledge (due to the use of GMM) even before training starts. Therefore, the performance after 50 epochs is almost same as after after 1 epoch in most of the cases. Most of the learning takes place during the preprocessing stage and not in the neural network. Thus, at last, we train a model without using GMM preprocessing. We use Wasserstein GANs with Gradient Penalty (WGAN-GP)~\cite{gulrajaniWGANGP2017,arjovskyWGAN2017}. Fig.~\ref{fig:synthetic-d} shows the results for WGAN-GP. This gives comparatively stable performance and shows gradual improvement with training in all datasets. Despite showing improvement in the quality of data that it generates with increasing number of epochs, it fails to overcome the overall scores achieved by CTGAN. This shows the importance of GMM preprocessing in obtaining useful synthetic data.

\subsubsection{\textbf{TVAE}}
In addition to the GAN based methods, we also use a variational auto-encoder (VAE) based method named tabular VAE (TVAE)~\cite{kingmaauto,xu2019modeling} to evaluate the synthesized tabular data. The results of TVAE on all 4 datasets is shown in Table~\ref{tab:results}. It can be observed that TVAE performs quite well (better than all other methods) on \textit{concrete data} but obtain average to poor performance on other three datasets. More detailed results are discussed in the Supplementary document.

\subsection{TabSynDex Score Comparison}
We compare the TabSynDex score of 8 existing tabular data generation methods and tabulate the same in Table~\ref{tab:results}. The TabSynDex score is computed on the test set of four datasets. Table~\ref{tab:results} shows the maximum similarities the various models attain over the datasets. Clearly, none of the models generate data with similarities anywhere near to those attained by real data subsets in Fig.~\ref{fig:real-a}. Also, as is visible from Table~\ref{tab:results}, the GAN based methods do not perform much better than standard statistical approaches like DataSynthesizer.

\subsection{Proposed and Existing Metric Score Comparison}
We compare TabSynDex score with the existing metric scores in Table~\ref{tab:metrics_dataset_comparison}. The metric scores for TVAE~\cite{kingmaauto,xu2019modeling} are shown for all 4 datasets. The \textit{JS-divergence}~\cite{zhao2022ctab} denotes the average JS-divergence between the corresponding categorical columns of real and synthetic data. The \textit{Wasserstein distance}~\cite{zhao2022ctab} denotes the average Wasserstein distance between the corresponding continuous columns of real and synthetic data. The \textit{correlation difference}~\cite{zhao2022ctab} denotes the L2-distance between correlation matrices of real and synthetic data. The~\textit{ML-Real} denotes the performance on the test set after training a machine learning model on real data and \textit{ML-Syn} denotes the same after training on synthetic data. The ML model performance is in terms of mean absolute error for all datasets except wine quality data in which f1-score is used. The~\textit{pMSE}~\cite{pmse_expected} is a measure of indistinguishability of synthetic data from real data. The \textit{TabSynDex} is the proposed metric. From Table~\ref{tab:metrics_dataset_comparison}, it can be observed that except~\textit{correlation difference} and ~\textit{TabSynDex}, all other metrics are unbounded. Therefore, it is not possible to interpret these metrics without a reference point. The \textit{correlation difference} does not cover many of desirable attributes of synthetic data as shown in Table~\ref{tab:metric_aspect_comparison}. Thus, TabSynDex is the only metric which can be used to measure the quality synthetic data on an absolute basis and without the need of a reference point.

\begin{table}[]
    \centering
    \resizebox{\columnwidth}{!}{
    \begin{tabular}{c|c|c|c|c}
    \hline
        {} & {Concrete~\cite{concrete_data}} & {Wine Quality~\cite{wine_data}} & {Power Plant~\cite{electrical_data}} & {News~\cite{news_data}}\\
        \hline
        {JS-Divergence~\cite{support_coverage, zhao2022ctab}} & {NA} & {0.17} & {NA} & {0.15} \\
        {Correlation Difference \cite{zhao2022ctab}} & {0.67} & {1.17} & {1.08} & {4.6} \\
        {Wasserstein Distance \cite{zhao2022ctab}} & {0.02} & {0.01} & {0.02} & {0.02} \\
        {ML-Real \cite{xu2018synthesizing, xu2019modeling}} & {7.33} & {0.99} & {0.121} & {3034.35} \\
        {ML-Syn \cite{xu2018synthesizing, xu2019modeling}} & {8.7} & {0.85} & {0.124} & {2469.91} \\
        {pMSE \cite{pmse_expected}} & {4.23} & {79.89} & {343.77} & {2402.8} \\
        {TabSynDex} & {0.72} & {0.45} & {0.65} & {0.40} \\
        \hline
    \end{tabular}
    }
    \caption{Scores by different evaluation metrics for synthetic data generated by TVAE model. \textit{NA: There were no categorical columns in the data}}
    \label{tab:metrics_dataset_comparison}
\end{table}

%Tabular data synthesis plays an important role in ensuring privacy preservation in the data-driven systems. 
\section{Conclusion}
The existing tabular data synthesis methods have shown promise in generating useful synthetic data. The generative adversarial networks (GAN) and variational auto-encoder (VAE) based methods have outperformed the traditional statistical methods. However, the lack of a uniform metric has caused inconsistency in the comparative evaluation of different methods. The existing metrics also do not have an implicit bound. This work presents a single score bounded universal metric TabSynDex for robust evaluation of synthetic tabular data. The proposed TabSynDex consists of five component scores capturing various desired qualities of synthetic data. It can be used as a universal metric to compare various GANs, VAEs, or other statistical methods for generating synthetic tabular data. We also show how this metric can be used to evaluate the training of GANs and VAEs to help discover new insights. The proposed metric is compared with the existing metrics based on several desired attributes such as inter-column correlation, distribution comparison, rare class coverage, implicit bound, and suitability to machine learning applications. In comparison with the existing metrics that cover few of these aspects, the TabSynDex covers all these attributes. In the experiments, it is observed that most of the existing tabular GAN and VAE methods are not competent for robustly synthesizing real-world data. The standard WGAN-GP and TVAE are relatively stable methods where all the component scores usually improve during training with some rare exceptions. Our work highlights the need for further research on the development of better tabular data synthesis methods. Another future work could be to study the relationship between the five constituent metrics and exploring the effect of using different weighing factor for each metric.
%Except TVAE for concrete dataset, none of the existing methods could achieve TabSynDex score even near to what real data subsets of equal size achieve. TabSynDex, equal weights are assigned to each of the five constituent metrics. However, different weights can be assigned to individual metrics and the assignment of weights will depend on the specific requirements for an application. 
\bibliographystyle{IEEEtran}
\bibliography{IEEEfull}

\clearpage
\iffalse
\fi
\end{document}

\makeatletter
\newcommand{\printfnsymbol}[1]{%
  \textsuperscript{\@fnsymbol{#1}}%
}
\makeatother

%title ideas: 
\title{Supplementary: A Universal Metric for Robust Evaluation of Synthetic Tabular Data}

%\author{Vikram S Chundawat\printfnsymbol{1}\thanks{$^*$Equal contribution.}, Ayush K Tarun\printfnsymbol{1}, Murari Mandal, Mukund Lahoti, Pratik Narang
%\thanks{Vikram S Chundawat, Ayush K Tarun, Mukund Lahoti, and Pratik Narang are with BITS Pilani, India (e-mail: \{f20180128, f20180258, pratik.narang, mukund.lahoti\}@pilani.bits-pilani.ac.in)}
%\thanks{Murari Mandal is with School of Computer Engineering, Kalinga Institute of Industrial Technology (Deemed to be University) Bhubaneswar, India (e-mail: murari.mandalfcs@kiit.ac.in)}
%}
\maketitle

\begin{figure*}
\centering
\begin{minipage}{\linewidth}
  \centering
  \includegraphics[width=0.7\linewidth, height=0.5cm]{legends.png}
  \begin{tabular}{ccc}
  \includegraphics[width=.20\linewidth]{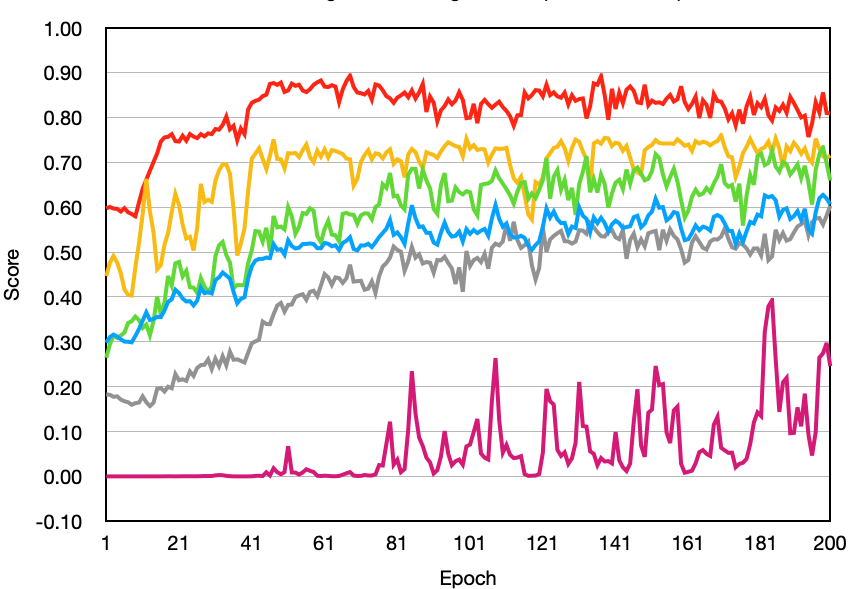}
    & \includegraphics[width=.20\linewidth]{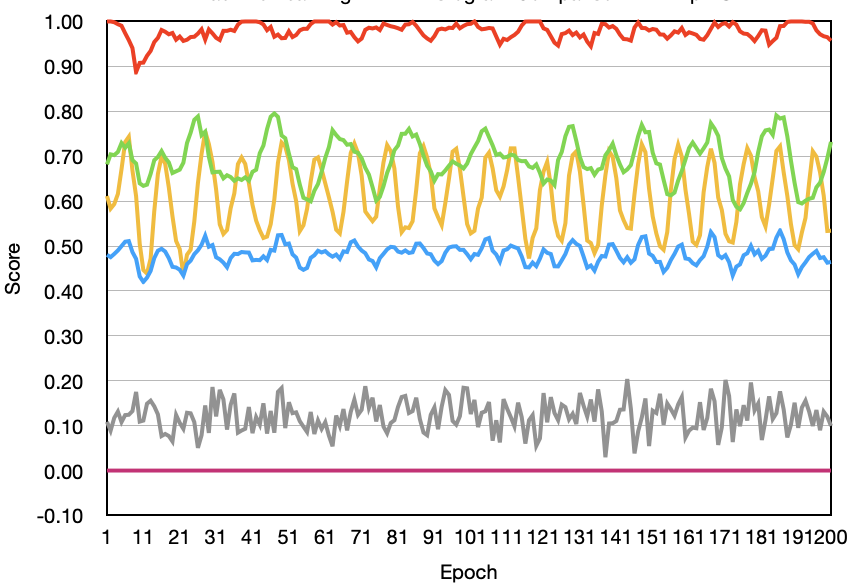}
    & \includegraphics[width=.20\linewidth]{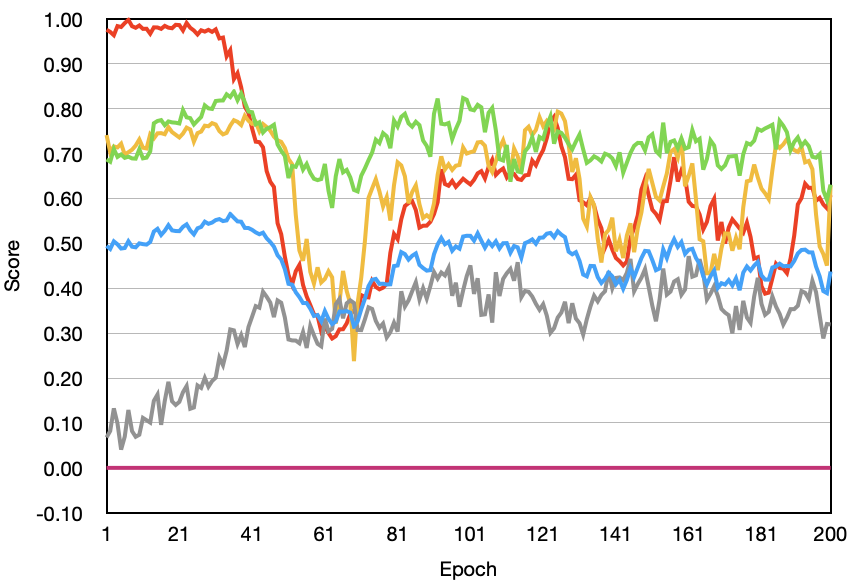}\\
  \scriptsize{i. WGAN-GP} & \scriptsize{ii. CTGAN} & \scriptsize{iii. TGAN}\\  
  \end{tabular}
  \vspace{-1em}
  %\subcaption{Training WGAN-GP, CTGAN, TGAN on the concrete data for 200 Epochs}
  %\label{fig:200_epoch_a}
\end{minipage}\par\bigskip
  
\begin{minipage}{\linewidth}
 \centering
\begin{tabular}{ccc}
  \includegraphics[width=.20\linewidth]{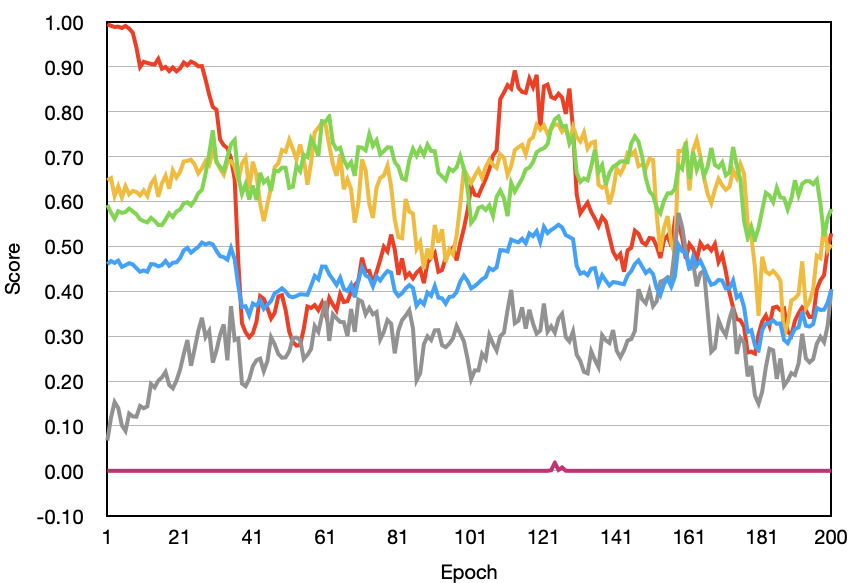}
    & \includegraphics[width=.20\linewidth]{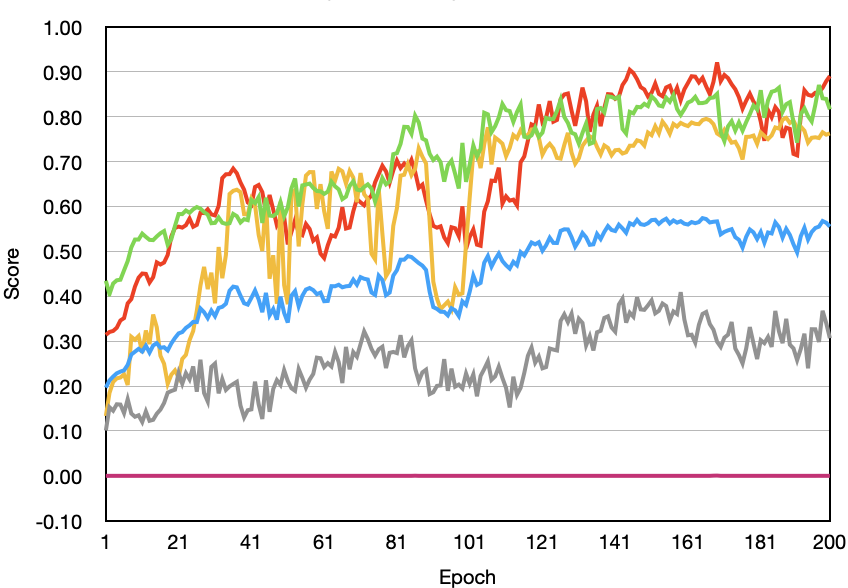}
    & \includegraphics[width=.20\linewidth]{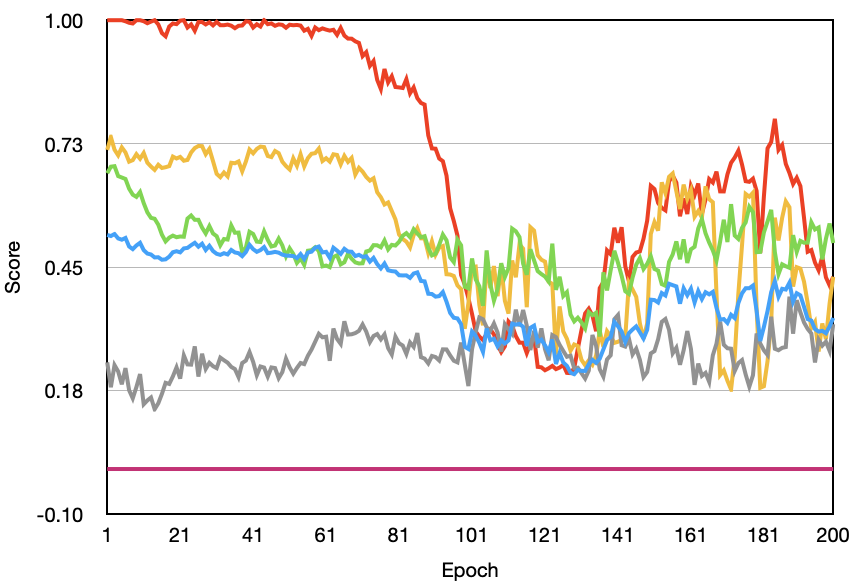}\\
  \scriptsize{iv. TGAN+GRU} & \scriptsize{v. Bidirectional TGAN} & \scriptsize{vi. VGMM+TGAN}\\
  \end{tabular}
  %\subcaption{Training TGAN+GRU, Bidirectional TGAN, VGMM+TGAN  on the concrete data for 200 Epochs}
  %\label{fig:200_epoch_b}
  \end{minipage}
%  \begin{minipage}{\linewidth}
% \centering
%\begin{tabular}{c}
%  \includegraphics[width=.23\linewidth]{img/TGAN_GRU_GMM_concrete_200_epochs_2.png}\\
%  \scriptsize{vii. VGMM+Bidirectional TGAN}\\
%  \end{tabular}
  %\subcaption{Training VGMM+Bidirectional TGAN on the concrete data for 200 Epochs}
%  \label{fig:200_epoch_c}
%  \end{minipage}
 \caption{The figure shows the effect of training for longer period. The training progression of WGAN-GP, CTGAN, TGAN, TGAN+GRU, Bidirectional TGAN, and VGMM+TGAN on the concrete data for 200 epochs is depicted.}
 \vspace{-1em}
\label{fig:200_Epochs_main}
\end{figure*}

\begin{figure}[]
\centering
\begin{minipage}{\linewidth}
  \centering
  \includegraphics[width=0.9\linewidth]{legends.png}
  \begin{tabular}{cc}
  \includegraphics[width=.45\linewidth]{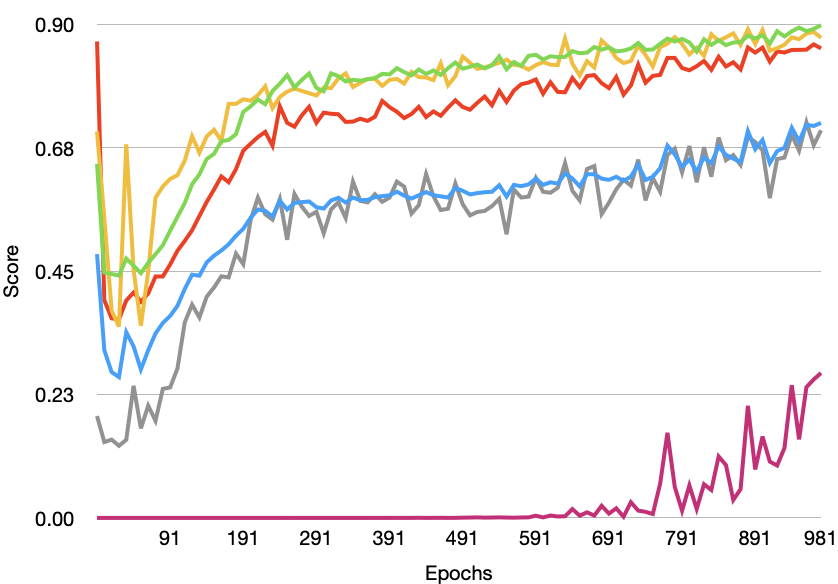}
    & \includegraphics[width=.45\linewidth]{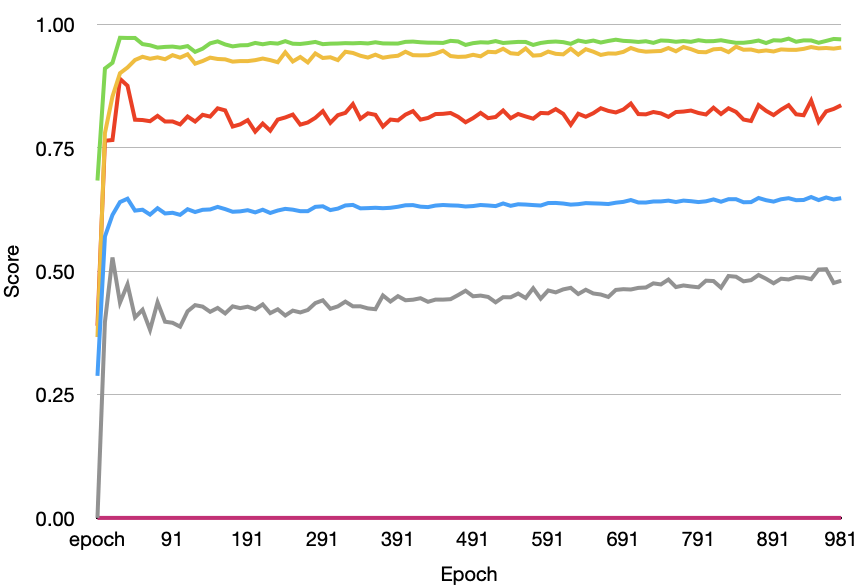}\\
      \scriptsize{i. Concrete Data} & \scriptsize{ii. Electrical Power Plant Data}\\
  \end{tabular}
  \end{minipage}
\caption{TabSynDex score of synthetic tabular data generated by TVAE method on \textit{concrete} and \textit{electrical power plant data}.}
\vspace{-1em}
\label{fig:tvae-results}
\end{figure}

\section{Additional Results on TVAE}
We use a variational auto-encoder (VAE) based method named tabular VAE (TVAE)~\cite{kingmaauto,xu2019modeling} to evaluate the synthesized tabular data. The results on \textit{concrete} and \textit{electrical power plant data} are shown in Fig.~\ref{fig:tvae-results}.  
In~\textit{concrete data}, the scores for basic statistics $S_{basic}$, correlation score $S_{corr_l}$, support coverage $S_{cr}$, and ML efficacy $S_{ml}$ increase gradually with increasing epochs of training. The $S_{pMSE}$ starts increasing only after 600 epochs. Similar behaviour is observed in \textit{electrical power plant data} except that the $S_{pMSE}$ remains flat throughout.

\section{Effect of Longer Training on the Model Performance}
We investigate the effect of training more number of epochs on the model performance. We train the models WGAN-GP, CTGAN, TGAN, TGAN+GRU, Bidirectional TGAN, and VGMM+TGAN for 200 epochs on the \textit{concrete data}. The training progression graphs are depicted in Fig.~\ref{fig:200_Epochs_main}. In most cases, the metrics (such as histogram comparison score $S_{cr}$, correlation score $S_{corr_l}$, etc.) obtained after 200 epochs are similar to what was obtained after 50 epochs. In TGAN and TGAN+GRU, all the component metrics of TabSynDex show a good degree of stability between 50-200 epochs. In Bidirectional TGAN, there is gradual improvement with increase in epochs in all the component metrics of TabSynDex, except $S_{pMSE}$ score. In VGMM+TGAN, there is some dip in the performance at 100-140 epochs. However, all the metrics recover to a good extent by the end of 200 epoch. Interestingly, we observe improvements in $S_{pMSE}$ score in WGAN-GP model. This is the only method that obtains some useful $S_{pMSE}$ score after longer training. Other methods do not improve in $S_{pMSE}$ score even after 200 epochs.

\section{Limitation}
Purdam and Elliot~\cite{purdam2007case} have shown that ensuring separability of the synthetic data from original data samples leads to a direct reduction in synthetic data utility. It implies that the utility of synthetic data and privacy of data are counter-intuitive measures. Therefore, aggregating both in a single metric to evaluate the tabular synthetic data generation models is not appropriate. To increase the utility of the synthetic data, an ideal metric should check how close the synthetic data is to the real data distribution. However, it does not ensure how much of the real data may be revealed (directly or indirectly) by the synthetic data. Some metrics such as \textit{membership disclosure} and \textit{attribute disclosure}~\cite{choi2017generating} are used to check the presence of a data point in the training data. Including such disclosure measures to the TabSynDex score is an interesting future scope of this work.

\bibliographystyle{IEEEtran}
\bibliography{IEEEfull}
\clearpage
\iffalse
\fi